%% file: acl_latex.tex
\documentclass[11pt]{article}
\usepackage[preprint]{acl}

\usepackage{times}
\usepackage{latexsym}
\usepackage[T1]{fontenc}
\usepackage[utf8]{inputenc}
\usepackage{microtype}
\usepackage{inconsolata}
\usepackage{graphicx}

\usepackage{microtype}
\usepackage{subcaption}

\usepackage{hyperref}
\usepackage{url}
\usepackage{graphicx}
\usepackage{booktabs}
\usepackage{wrapfig}
\usepackage{enumitem}
\usepackage{capt-of}
\usepackage{tabularx}
\usepackage{ragged2e}
\usepackage{tikz}
\usepackage{bm}
\usepackage[normalem]{ulem}
\usepackage{colortbl}
\usepackage{xcolor}
\usepackage{float}

\usepackage{booktabs}
\usepackage{xcolor}
\usepackage{amssymb}
\usepackage{multirow}
\usepackage{arydshln}
\definecolor{posgreen}{HTML}{2E8B57}
\definecolor{negred}{HTML}{B22222}
\newcommand{\up}{\textcolor{posgreen}{$\blacktriangle$}}
\newcommand{\down}{\textcolor{negred}{$\blacktriangledown$}}

\usepackage{amsmath}
\usepackage{amssymb}
\usepackage{mathtools}
\usepackage{amsthm}

\usepackage[capitalize,noabbrev]{cleveref}

\usepackage[textsize=tiny]{todonotes}

\usepackage{tcolorbox}
\tcbuselibrary{skins}
\definecolor{BlueViolet}{cmyk}{0.86,0.91,0,0.04}

\makeatletter
\newcommand{\storeabstract}[1]{\gdef\@storedabstract{#1}}
\def\@storedabstract{}

\def\@metadatalist{}

\newcommand{\metadata}[2]{%
    \ifx\@metadatalist\empty
        \gdef\@metadatalist{\textbf{#1:} #2}%
    \else
        \g@addto@macro\@metadatalist{\\[0.2em] \textbf{#1:} #2}%
    \fi
}

\AtBeginDocument{%
\def\@maketitle{%
    \setlength{\fboxrule}{1pt}%
    \setlength{\fboxsep}{1em}%
    \begin{tcolorbox}[
        width=\textwidth,
        colback=blue!2,
        colframe=white,
        boxrule=0pt,
        arc=6pt,
        outer arc=6pt,
        boxsep=0em,
        enhanced,
        frame style={draw=blue!20, line width=1pt, dashed}
    ]
        \centering
        \vskip 0.4em
        {\LARGE\sc \@title\par}
        \vskip 0.9em
        {\normalsize \@author\par}
        \ifx\@storedabstract\empty\else
            \vskip 1.2em
            \begin{justify}
            \textbf{Abstract:} \@storedabstract
            \end{justify}
        \fi
        \ifx\@metadatalist\empty\else
            \vskip 0.4em
            {\raggedright\@metadatalist\par}
        \fi
        \vskip 0.4em
    \end{tcolorbox}
    \vskip 1.2em
}%
}
\makeatother

\title{\textbf{Scale or Reason?}\\ \Large A Compute-Equivalent Analysis of Reasoning Distillation}

\author{\textbf{Nicolas Boizard\textsuperscript{1,5}} \quad
\textbf{Hippolyte Gisserot-Boukhlef\textsuperscript{2,5}} \\ [0.3em]
\textmd{Kevin El Haddad\textsuperscript{1,4} \quad
Céline Hudelot\textsuperscript{5}} \quad
Pierre Colombo\textsuperscript{3}\\[0.6em]
\textsuperscript{1}Diabolocom \quad
\textsuperscript{2}Artefact Research Center \quad
\textsuperscript{3}Cohere \quad
\textsuperscript{4}ISIA Lab, University of Mons \\ [0.2em]
\textsuperscript{5}MICS, CentraleSupélec, Université Paris-Saclay}

\storeabstract{Distilling reasoning traces from strong teacher models has become the standard recipe for building capable small language models. Yet reasoning traces are 5-20$\times$ longer than standard instruction fine-tuning (IFT) outputs, meaning every practitioner who chooses reasoning distillation implicitly forgoes training a larger IFT model on the same compute budget. Whether this trade-off is worthwhile remains unaddressed. We study it with a controlled experiment: a single teacher generates paired IFT and reasoning outputs for identical prompts by toggling only its reasoning mode, isolating supervision format as the sole variable. Training students at five scales (0.5B to 14B) and evaluating on 18 benchmarks, we find that at matched FLOPs, IFT lies on or near the Pareto frontier across the majority of configurations. Reasoning reaches the Pareto frontier only on open-ended tasks at 7B and above. Even there, a sequential curriculum mixing just 25-50\% reasoning data with IFT captures most of the accuracy benefit at far lower compute cost.}

\metadata{Correspondence}{\textcolor{BlueViolet}{nicolas.boizard@centralesupelec.fr}}
\metadata{Website}{\textcolor{BlueViolet}{https://huggingface.co/scale-or-reason}}
\metadata{Date}{September 26, 2025}

\begin{document}
\maketitle

\section{Introduction}
\label{sec:introduction}

\begin{figure}[t!]
    \centering
    \includegraphics[width=0.45\textwidth]{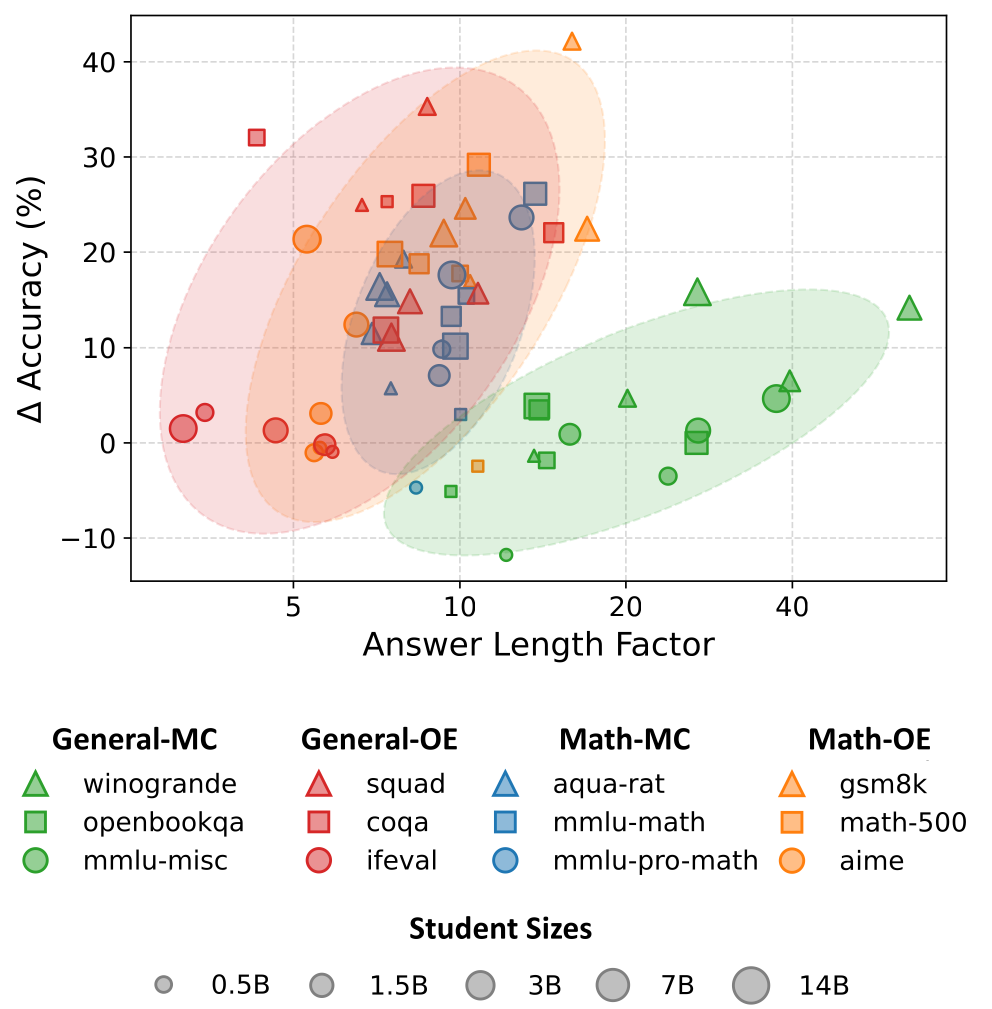}
    \caption{Compute overhead of reasoning distillation. Each point is a benchmark-model-size pair. X-axis: extra-token factor when switching from IFT to reasoning supervision. Y-axis: accuracy gain (\%). The overhead is large (5-20$\times$) and heterogeneous across tasks.}
    \label{fig:task_characterization}
\end{figure}

Reasoning distillation has become the dominant training recipe for capable small language models \citep{deepseekai2025deepseekr1, qwen3, mistralai2025magistral}. Yet every reasoning trace is 5-20$\times$ longer than a standard IFT output (\autoref{fig:task_characterization}), and this length gap has a direct cost: training on reasoning data consumes proportionally more compute. A practitioner with a fixed budget who chooses reasoning distillation implicitly gives up the option of training a larger IFT model instead, an alternative that standard scaling laws suggest is a highly efficient use of additional compute \citep{kaplan2020scaling, chinchilla}. Whether reasoning justifies this cost has not been studied: DeepSeek-R1 \citep{deepseekai2025deepseekr1}, s1 \citep{muennighoff2025s1simpletesttimescaling}, and \citet{snell2024scaling} all evaluate at fixed model sizes, leaving the compute-allocation question entirely open. This leads directly to our question:

\vspace{0.2cm}
\begin{center}
\noindent\fbox{%
    \parbox{0.95\linewidth}{%
        \centering
        \emph{For a given compute budget, should a practitioner invest in reasoning distillation, or train a larger IFT model?}
    }%
}
\end{center}
\vspace{0.2cm}

\vspace{0.4cm}

To isolate the effect of supervision format from all other variables, we use a single teacher that generates paired IFT and reasoning outputs for identical prompts by toggling its reasoning mode on and off. Both formats are thus distilled from the same teacher weights in a controlled setup, ensuring that supervision format is the sole variable. We train \texttt{Qwen2.5} students at five scales (0.5B to 14B), track FLOPs precisely across all configurations, and evaluate on 18 benchmarks spanning four task families.  We replicate all key findings using a second teacher-student pair (\texttt{Nemotron-Super-49B} with \texttt{Gemma-3}; \autoref{apd:other_teacher_student}), confirming that results are not specific to a single architecture or teacher. This yields three findings:

\begin{itemize}[leftmargin=1.2em,itemsep=2pt,topsep=2pt]
    \item \textbf{Scaling IFT is compute-efficient by default.}
    At matched training and inference FLOPs, IFT models lie on or near the Pareto frontier across the majority of configurations. Reasoning only approaches the frontier on open-ended tasks and at larger scales (\S\ref{sec:scale_or_reason}).

    \item \textbf{Format and capacity jointly determine the value of reasoning.}
    Output format, rather than knowledge domain, is the primary predictor: open-ended tasks consistently benefit from reasoning distillation, whereas multiple-choice tasks do not. Model capacity acts as a secondary prerequisite, as reasoning captures the Pareto frontier only at 7B parameters and above (\S\ref{sec:analysis}).

    \item \textbf{A little reasoning goes a long way.}
    In the regime where reasoning is beneficial, a sequential curriculum using just 25 to 50\% reasoning data captures most of the accuracy benefit at a fraction of the training cost. Furthermore, mixed training below the 50\% threshold delivers enhanced accuracy while strictly maintaining efficient inference costs (\S\ref{sec:budget_reasoning}).
\end{itemize}

Together, these findings provide a systematic answer to the compute allocation question in reasoning distillation.

\section{Experimental Setup}
\label{sec:experimental_framework}

\begin{table*}[t]
\centering
\renewcommand{\arraystretch}{1.1}
\resizebox{\textwidth}{!}{
\input{tables/results_mono_phasic}
}
\vspace{2mm}
\caption{Average scores for IFT versus reasoning distillation by model size, corpus, and task format. $\Delta$ denotes absolute accuracy shifts from standard IFT. Accuracy breakdowns by dataset are detailed in \autoref{apd:task_level_results}.}
\label{tab:results_mono_phasic}
\end{table*}

\subsection{Distillation}

\paragraph{Controlled paired design.}
Our central methodological challenge is to compare IFT and reasoning supervision without confounding label quality with supervision format. We address this by using a single teacher, \texttt{Qwen3-235B-A22B} \citep{qwen3}, to generate both formats for every prompt: IFT outputs are produced with reasoning mode disabled, reasoning outputs with it enabled. Student models in the IFT condition are trained on $\langle \text{question}, \text{answer} \rangle$ pairs; in the reasoning condition on $\langle \text{question}, \text{trace}, \text{answer} \rangle$ tuples. Because both formats come from the same teacher weights and the same prompts, supervision format is the sole independent variable.

\paragraph{Students.} We train \texttt{Qwen2.5} base models \citep{qwen2.5} at five parameter scales: 0.5B, 1.5B, 3B, 7B, and 14B. Students are drawn from a different model family than the teacher to limit shared pretraining overlap. Training data covers two domains: a general-purpose corpus of 1.3M pairs from \texttt{Infinity-Instruct} \citep{infinityinstruct}, and a reasoning-centric corpus of 300K pairs from the math subset of the \texttt{Llama-Nemotron-Post-Training-Dataset} \citep{nemotron}.

\paragraph{Training.} We study sequential training across reasoning proportions $\rho \in \{0\%, 25\%, 50\%, 75\%, 100\%\}$, where $\rho$ is the percentage of reasoning instances in the dataset. At $\rho = 0\%$ the model sees only IFT supervision; at $\rho = 100\%$ only reasoning supervision. Intermediate values correspond to a curriculum where IFT examples come first, followed by reasoning examples. We also evaluate a mixed training baseline in which IFT and reasoning data are shuffled together within a single phase (\S\ref{sec:budget_reasoning}). All models train for one epoch on 16 H100 GPUs with a fixed global batch size of 32; learning rates are selected by grid search (\autoref{apd:hyperparameters}).

\subsection{Evaluation}

\paragraph{Tasks.} Our evaluation centers on 18 benchmarks structured along two dimensions: domain (general vs.\ math) and answer format (multiple-choice (MC) vs.\ open-ended (OE)). This taxonomy defines four task families: General-MC (\texttt{winogrande}~\citep{winogrande}, \texttt{openbookqa}~\citep{openbookqa}, \texttt{mmlu-misc}~\citep{mmlu}), General-OE (\texttt{squad}~\citep{squad}, \texttt{coqa}~\citep{coqa}, \texttt{ifeval}~\citep{ifeval}), Math-MC (\texttt{aqua-rat}~\citep{aquarat}, \texttt{mmlu-math}~\citep{mmlu}, \texttt{mmlu-pro-math}~\citep{mmlupro}), and Math-OE (\texttt{gsm8k}~\citep{gsm8k}, \texttt{math-500}~\citep{math500}, \texttt{aime}~\citep{aime}). Additional tasks are introduced selectively in \S\ref{sec:analysis} to address specific experimental needs.

\paragraph{LLM-as-a-Judge.} Answers are scored by a majority-vote jury of three judge models: \texttt{Llama-3.3-70B-Instruct} \citep{grattafiori2024llama3herdmodels}, \texttt{Nemotron-Ultra-253B-v1} \citep{nemotron}, and \texttt{GPT-OSS-120B} \citep{openai2025gptoss120bgptoss20bmodel}. Each judge follows a two-step protocol: first extracting the final answer from the model output while ignoring any intermediate reasoning, then assessing its semantic equivalence to the reference. This ensures that scores reflect answer correctness, not trace quality, and prevents reasoning-trained models from receiving credit for verbose but wrong answers. Median absolute error across judges is 1.2 points; pairwise agreement is below 2 points on all benchmarks. Distilled models are evaluated zero-shot while base checkpoints rely on three-shot prompting. Full prompts and judge statistics are provided in \autoref{apd:prompting_details}.

\paragraph{Compute cost.} Because reasoning traces are 5-20$\times$ longer than IFT outputs, reasoning training and inference consume proportionally more compute for both forward and backward passes. We systematically track this overhead in FLOPs and report the formula used in \autoref{apd:flops_computation}.
This disparity is the foundation for the compute-matched comparisons in \S\ref{sec:scale_or_reason}.

\begin{figure*}[t]
\centering
\includegraphics[width=\textwidth]{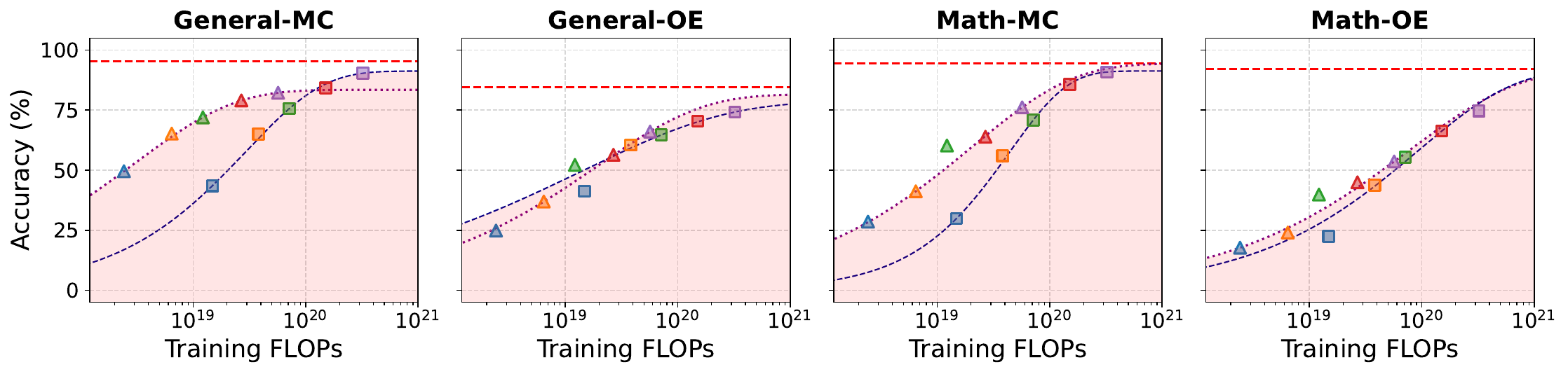}
\includegraphics[width=\textwidth]{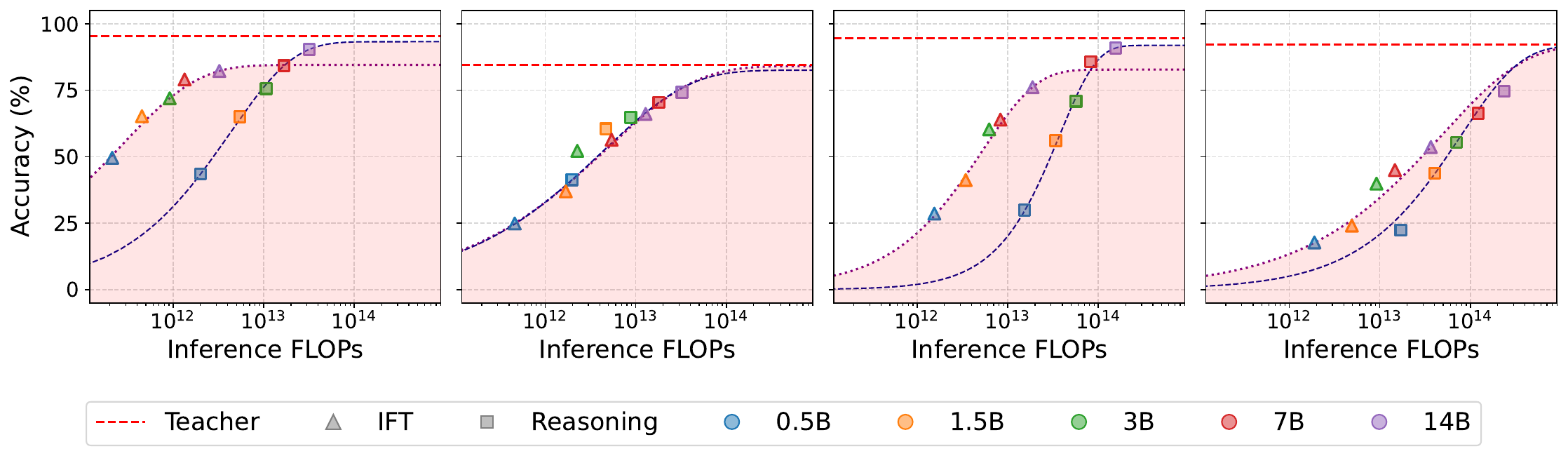}
\caption{Accuracy as a function of training FLOPs (top) and inference FLOPs (bottom) for IFT and reasoning models on the general-purpose corpus. Each point corresponds to one model size; curves are saturating growth fits to the observed data (see \autoref{apd:flops_computation} for details).}
\label{fig:general_pareto}
\end{figure*}

\section{Is Reasoning Worth the Compute Cost?}
\label{sec:scale_or_reason}
We approach this question in three distinct steps: comparing the two supervision formats at equal model size, at matched training compute, and at matched inference compute.

\subsection{At Fixed Size, Reasoning Helps}
\label{sec:fixed_size}

To establish an initial baseline, \autoref{tab:results_mono_phasic} presents the raw performance of both supervision paradigms across all model sizes and the two relevant training corpora: the general-purpose one derived from \texttt{Infinity-Instruct} and the math-centric one derived from \texttt{Llama-Nemotron-Post-Training-Dataset}.

\paragraph{General-purpose training.}
Reasoning supervision consistently outperforms IFT when trained on the general-purpose corpus, with the largest gains on open-ended and mathematical tasks. On Math-OE, a 3B reasoning model already surpasses the 14B IFT baseline outright (55.4 vs.\ 53.7); on General-OE, it comes within 1.4 points of it (64.7 vs.\ 66.1). On both MC families, the 14B reasoning model narrows the performance gap with the teacher to within 4-5 points, despite being roughly 17$\times$ smaller. The pattern holds across nearly all conditions with one notable exception: on General-MC, IFT leads at 0.5B, and the two formats remain effectively tied at 1.5B. However, from 3B onward, reasoning re-establishes a definitive lead.

\paragraph{Math-centric training.}
When trained on the math-centric corpus, reasoning supervision again outperforms IFT from 1.5B onward on math tasks. More strikingly, from 3B upward, math-centric reasoning models surpass even general-purpose reasoning on both Math-OE (57.5 vs.\ 55.4) and Math-MC (72.2 vs.\ 70.9), despite training on just 300K samples compared to 1.3M. However, this specialization comes at a cost: on general-domain tasks, math-centric reasoning lags well behind general-purpose reasoning at smaller scales, and the gap closes only partially by 14B (General-MC: 90.3 vs.\ 90.4; General-OE: 69.2 vs.\ 74.2). Conversely, standard IFT models trained on the exact same math-centric corpus do not exhibit this penalty, remaining competitive on general-domain tasks across all tested scales. This demonstrates that the general-task degradation is tied to the reasoning format rather than the math-centric data itself. This penalty is particularly severe at smaller scales, demonstrating that distilling general capabilities from a reasoning-intensive domain (e.g., mathematical problem solving or code generation, as commonly used in RL pipelines) strictly requires larger parameter counts. Consequently, we report the remainder of our experiments with the general-purpose training corpus.

\paragraph{The limits of parameter-matched comparisons.}
Both supervision formats improve with model size, but at different rates. IFT improves steadily across all four task families, exhibiting consistent gains at each size step; yet even at 14B, it remains well below teacher accuracy, with the largest gap appearing on open-ended tasks. Reasoning scales more steeply and outperforms IFT in 70\% of scale-corpus configurations (28 out of 40 pairs), though not uniformly. While these results make a strong case for reasoning at a fixed parameter count, this comparison carries a fundamental limitation: equal model size does not imply equal compute. Because a 3B reasoning model consumes substantially more FLOPs than a 3B IFT model, this disparity changes the efficiency picture entirely.

\subsection{At Fixed FLOPs, IFT Dominates}
\label{sec:fixed_flops}

\begin{figure*}[t]
\centering
\includegraphics[width=\textwidth]{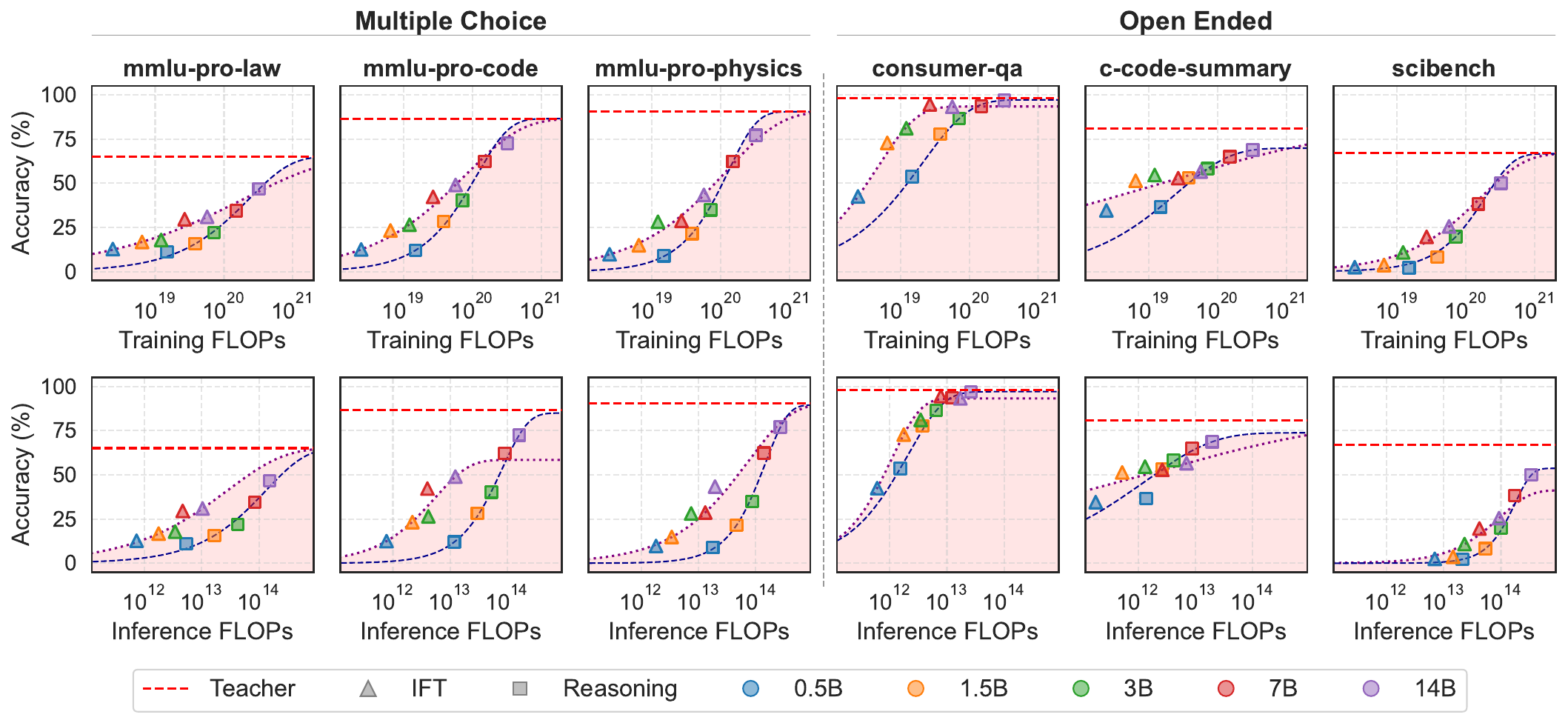}
\caption{Accuracy as a function of training FLOPs (top) and inference FLOPs (bottom) for IFT and reasoning models on reasoning-intensive tasks.}
\label{fig:reasoning_pareto}
\end{figure*}

In this section, we shift our focus to compute efficiency by plotting accuracy against training and inference FLOPs for IFT and reasoning distillation on the general-purpose corpus.

\paragraph{Training efficiency.}
Contrary to the fixed capacity comparison, when compute is held constant rather than model size, the efficiency picture reverses entirely. \autoref{fig:general_pareto} reveals that IFT models define the Pareto frontier of performance as a function of the training budget, with no reasoning configuration matching the accuracy of a compute-equivalent IFT model at any tested scale. This establishes that scaling model size is a more compute-efficient strategy for maximizing performance across the majority of our evaluations. For example, on general multiple-choice tasks, despite being twice the size, a 7B IFT model required 5$\times$ less training compute to achieve similar accuracy compared to a 3B reasoning-distilled model. While the cost overhead is smaller on open-ended tasks, a gap remains; for instance, on open-ended math tasks, a 7B IFT model achieves performance similar to a 1.5B reasoning model while being 1.6$\times$ less costly to train. However, we observe that the performance gap at matched training FLOPs continuously narrows with model scale, suggesting this disparity may close beyond 14B parameters.

\paragraph{Inference efficiency.}
Shifting focus from training to deployment, the pattern holds (\autoref{fig:general_pareto}). Because reasoning models generate outputs 7-15$\times$ longer than standard IFT models of the same size, their inference cost is substantially higher at every scale. On OE tasks, reasoning models still generate outputs approximately 7$\times$ longer than their compute-equivalent IFT counterparts, yet they approach or even join the Pareto frontier. On MC tasks, however, this overhead grows to ${\sim}$10-15$\times$, causing reasoning models to fall well behind IFT efficiency across most evaluated scales.


\section{What Drives Reasoning Efficiency?}
\label{sec:analysis}
\S\ref{sec:scale_or_reason} demonstrates that while standard IFT generally dominates at fixed compute, systematic exceptions emerge for open-ended tasks at larger parameter scales. This section investigates why output format and model capacity dictate when reasoning distillation justifies its computational overhead.

\subsection{Format Predicts Value; Domain Does Not}\label{sec:format_vs_domain}
Reasoning models incur significantly higher overhead on multiple-choice tasks than on open-ended tasks across diverse domains, questioning whether distillation efficiency is driven by domain complexity or output format.

\paragraph{Isolating format from domain.}
To disentangle these factors further, we evaluate three additional reasoning-intensive domains featuring one open-ended and one multiple-choice benchmark each: law (\texttt{mmlu\_pro\_law} \citep{mmlupro}, \texttt{consumer\_qa} \citep{guha2023legalbenchcollaborativelybuiltbenchmark}), code (\texttt{mmlu\_pro\_code}, \texttt{C\_Code\_Summarization} \citep{liu2021retrievalaugmentedgenerationcodesummarization}), and physics (\texttt{mmlu\_pro\_physics}, \texttt{scibench} \citep{wang2024scibenchevaluatingcollegelevelscientific}). Holding the knowledge domain constant while varying task format provides a precise test. If domain dictates efficiency, reasoning-intensive tasks will uniformly favor reasoning models. Conversely, if output format dominates, open-ended and multiple-choice versions of the same domain will diverge, with one capturing or resting near the Pareto frontier alongside IFT models, and the other falling behind.

\begin{figure*}[t]
\centering
\includegraphics[width=\textwidth]{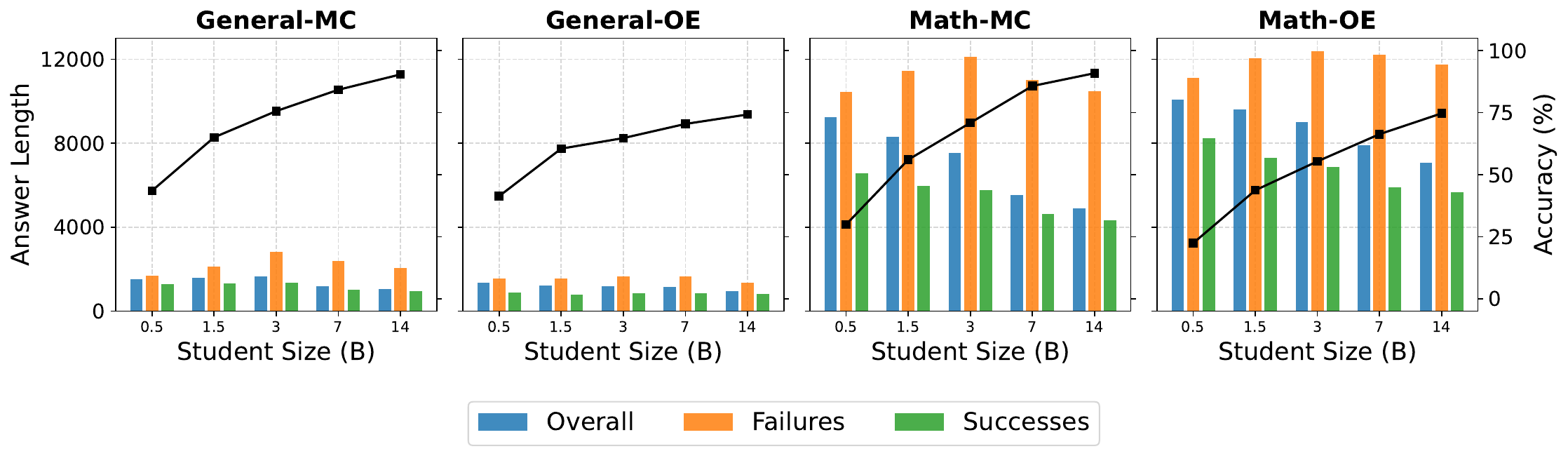}
\caption{Answer length analysis across student sizes and correctness in reasoning models.}
\label{fig:answer_length}
\end{figure*}

\paragraph{Output format dictates efficiency for reasoning-intensive tasks.} 
As initially illustrated in \autoref{fig:general_pareto} and further complemented by \autoref{fig:reasoning_pareto}, the two task formats exhibit a clear divergence: open-ended benchmarks place reasoning models on or near the Pareto frontier for both training and inference efficiency, while a significant gap remains for multiple-choice evaluations. For example, on \texttt{C\_Code\_Summarization}, a 7B reasoning model matches an IFT model at identical inference cost with a training overhead of only 1.3$\times$. Conversely, on multiple-choice benchmarks like \texttt{mmlu\_pro\_code}, a 10$\times$ inference compute gap exists between a 7B IFT model and a 3B reasoning model achieving similar performance, although this narrows with scale. These results confirm that while knowledge domain exerts marginal influence (math and reasoning-intensive multiple-choice tasks benefit slightly more than general ones), output format strictly dictates efficiency. For open-ended setups, this Pareto equivalence transforms the architectural decision into a practical exchange between model capacity and generation length, allowing practitioners to deploy smaller reasoning models for memory constraints, or larger IFT models to minimize latency.

\subsection{Model Capacity Determines Whether Traces Help or Hurt}
\label{sec:capacity_threshold}
While output format dictates where reasoning distillation is computationally viable, model capacity acts as a strict secondary prerequisite.

\paragraph{Reasoning dominates from 7B upward.} 
Beyond this capacity threshold, reasoning models reliably capture both the training and inference Pareto frontiers. On open-ended tasks (\autoref{fig:general_pareto}, \autoref{fig:reasoning_pareto}), reasoning models match or exceed IFT accuracy while maintaining optimal compute efficiency. On multiple-choice tasks, reasoning models at this scale surpass IFT in absolute accuracy, though they incur higher inference costs that deviate from established scaling trends.

\paragraph{Incorrect traces are longer, not deeper.} 
The relative difference in inference cost between consecutive reasoning model scales decreases steadily with model size (\autoref{fig:general_pareto}, \autoref{fig:reasoning_pareto}, and highlighted in \autoref{apd:inference_cost_scaling_trends}). Plotting the generation length of correct versus incorrect reasoning traces against model size and accuracy (\autoref{fig:answer_length}) reveals a consistent relationship: across all scales and task families, incorrect answers yield traces 1.5 to 2$\times$ longer than correct ones. Rather than reflecting deeper processing, longer traces correlate directly with failure. Qualitative inspection reveals smaller models frequently explore dead ends or restate problems without making tangible progress. As capacity increases, two phenomena emerge:

\begin{enumerate}[leftmargin=1.2em,itemsep=0.5pt,topsep=1pt]
    \item The length ratio between failures and correct answers shrinks (from roughly 2$\times$ at 0.5B to 1.5$\times$ at 14B), indicating larger models allocate generation effort more efficiently.
    \item Because accuracy inherently increases with scale, the average output length naturally gravitates toward the shorter, correct trace length.
\end{enumerate}

\noindent Conversely, average generation length for smaller models is skewed by long, incorrect traces, severely degrading the accuracy to inference cost ratio. This makes small reasoning models computationally expensive and practically ineffective. While this strong correlation makes trace length a natural confidence signal (similar to multi sampling strategies utilizing answer length as an uncertainty measure \citep{hassid2026dontoverthinkitpreferring}), attempting a simple early stopping strategy based on length thresholds proved unsuccessful. The reduction in inference FLOPs is entirely offset by a corresponding drop in accuracy without shifting the Pareto frontier, an outcome we consequently report in \autoref{apd:early_stopping}.

\begin{figure*}[t]
\centering
\includegraphics[width=\textwidth]{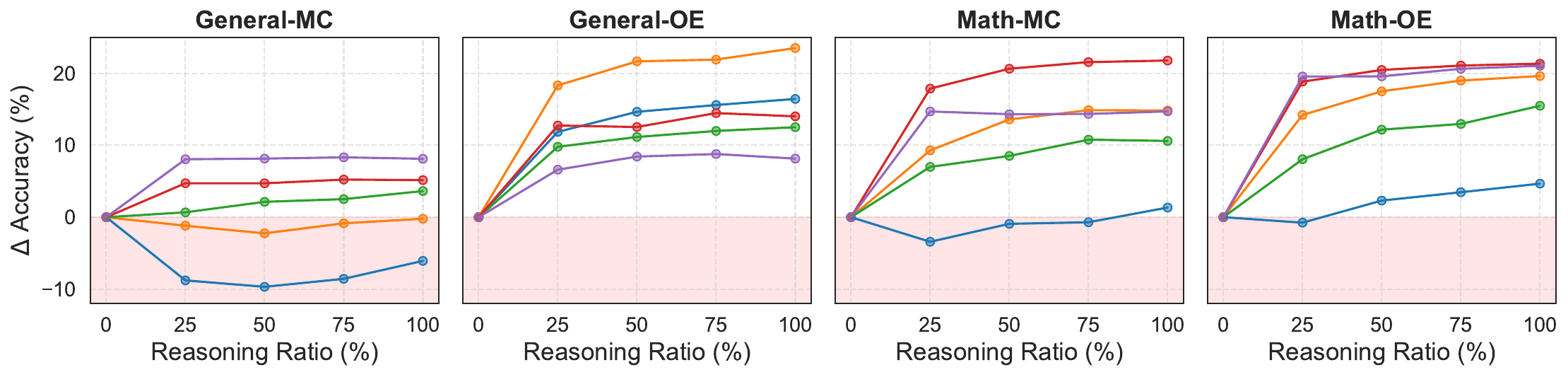}
\includegraphics[width=\textwidth]{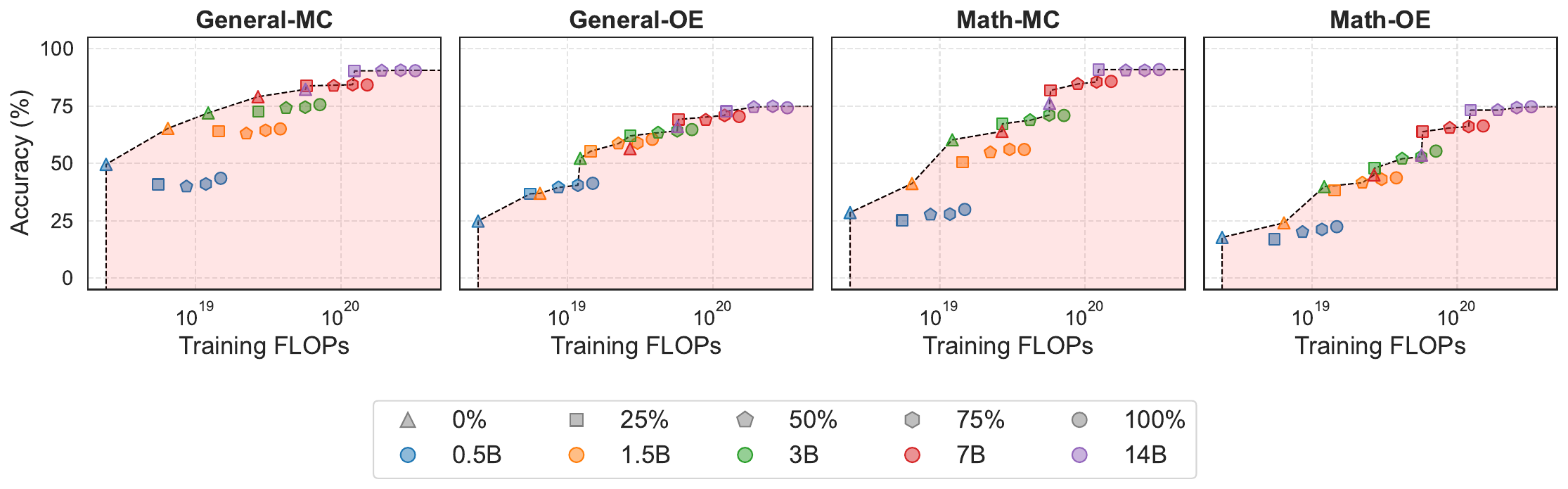}
\caption{Impact of the reasoning proportion on downstream performance in the sequential training scenario (standard IFT first, then reasoning). A 25\% reasoning proportion captures most gains on Math\_OE at a fraction of the compute. Results show the accuracy gap relative to the IFT baseline (0\%).}
\label{fig:reasoning_ratio}
\end{figure*}

\paragraph{Format and capacity jointly define the beneficial regime.} 
Together, output format and model capacity define when reasoning distillation earns its compute cost. open-ended formats make reasoning a viable alternative to IFT: both reach comparable accuracy on the efficiency frontier, reducing the choice to a deployment constraint. Scaling beyond 7B pushes this further: reasoning overtakes IFT on the frontier, becoming the dominant strategy for open-ended tasks.

\section{Leveraging IFT and Reasoning with Hybrid Data Mixtures} \label{sec:budget_reasoning}

\S\ref{sec:scale_or_reason} demonstrates that reasoning distillation boosts performance but carries a steep compute premium, with training and inference costs running 5 to 20$\times$ higher than standard IFT. \S\ref{sec:analysis} traced this overhead to smaller models generating inefficiently: producing long traces that correlate with failure rather than reasoning depth, driving up costs without proportionate accuracy gains. This section investigates whether this computational burden can be reduced. We study two hybrid strategies blending IFT- and reasoning-style data in varying proportions: \emph{sequential training} (IFT followed by reasoning) and \emph{mixed training} (both data types combined and shuffled in a single training pass).


\paragraph{Sequential training recovers most reasoning gains with minimal reasoning data.} Sequential training addresses a direct question: how much reasoning data is required? \autoref{fig:reasoning_ratio} demonstrates that performance gains emerge at just a 25\% reasoning ratio (meaning 75\% IFT data followed by 25\% reasoning traces) and remain broadly consistent across task families and model sizes, with the notable exception of the smallest models (0.5B and 1.5B) evaluated on general multiple-choice tasks. At every intermediate ratio (25\%, 50\%, and 75\%), sequential models successfully rest on the Pareto frontier for open-ended tasks. This matches the efficiency of full reasoning distillation at a mere fraction of its training cost. 

\begin{figure*}[t]
\centering
\includegraphics[width=\textwidth]{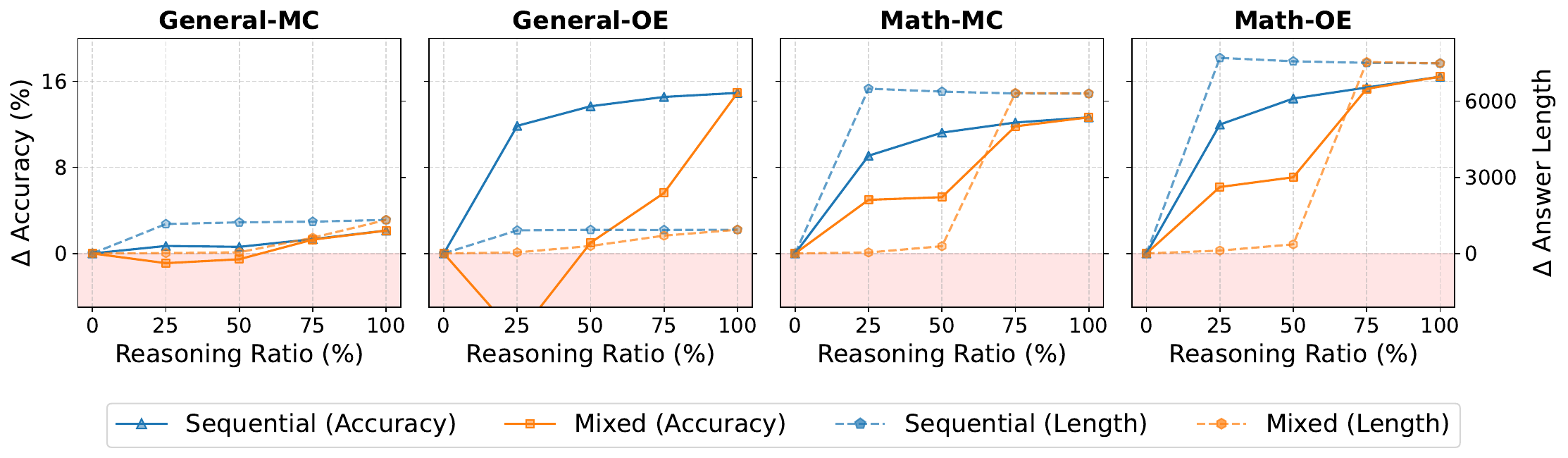}
\caption{Sequential vs mixed training across varying reasoning proportions. Solid lines indicate the accuracy gap relative to the IFT baseline, while dashed lines represent average answer length in tokens. All values are averaged across model sizes.}
\label{fig:mixed_training}
\end{figure*}

\paragraph{Mixed training enables reasoning improvements without increasing inference cost.} Mixed training pursues a distinctly different objective. Rather than strictly maximizing overall accuracy, it investigates whether a model can acquire reasoning capabilities while still retaining the concise output style typical of standard IFT. \autoref{fig:mixed_training} demonstrates that this is achievable below a specific data threshold. At reasoning proportions between 25 and 50\%, models improve substantially over the IFT baseline on mathematical tasks while maintaining short outputs. This allows them to gain accuracy without incurring the typical inference compute penalty.

\paragraph{Mixed training exhibits a phase transition once reasoning data becomes predominant, collapsing into purely reasoning-driven behavior.}
While mixed training remains beneficial at lower proportions of reasoning data, a clear mode switch emerges beyond the 50\% threshold. Models begin to default entirely to reasoning style generation at inference time, which completely erases any inference cost advantage. Output lengths converge toward full reasoning distillation, while accuracy remains slightly behind that of sequential training at the 75\% mark. Extending the benefits of mixed training beyond this point would likely require an explicit mechanism to control exactly when reasoning is invoked \citep{byrd2022predicting, alkhuzaey2024text}, a promising direction we leave for future work.

\paragraph{Sequential and mixed training serve complementary objectives.} Sequential training reduces the training compute required to approach full reasoning performance: utilizing just 25\% reasoning data recovers most accuracy gains at a fraction of the cost. Mixed training reduces inference cost: below the 50\% threshold, models improve over standard IFT while preserving output brevity. Together, these findings establish that committing fully to reasoning distillation is unnecessary. When training compute is the primary constraint, sequential mixing efficiently captures most performance gains; when inference cost is the constraint, mixed training below the 50\% threshold offers a practical middle ground.

\section{Related Work}
\label{sec:related_work}

\paragraph{Instruction tuning and reasoning.}  
Instruction Fine-Tuning (IFT) has been the standard recipe for aligning LLMs with human instructions \citep{wei2022finetuned, ouyang2022traininglanguagemodelsfollow, chung2022scalinginstructionfinetunedlanguagemodels}. Chain-of-Thought (CoT) extended this paradigm by supervising intermediate reasoning steps, yielding strong gains on arithmetic, symbolic, and commonsense reasoning benchmarks \citep{rajani2019explainyourselfleveraginglanguage, nye2021workscratchpadsintermediatecomputation, gsm8k, wei2022cot, kojima2022cot}.

\paragraph{Knowledge distillation.}  
Knowledge Distillation (KD) transfers capabilities from strong teachers to smaller students \citep{kdBuciluundefined, hinton2015distilling}. Beyond representation-based KD, text-based distillation has become central for reasoning: large teacher models generate either IFT- or reasoning-style traces that guide student learning \citep{kim-rush-2016-sequence, Zhou_2023, hsieh2023distillingstepbystepoutperforminglarger, he2024teacherlmteachingfishgiving}. This approach reduces the cost of expensive RL while preserving performance \citep{deepseekai2025deepseekr1, qwen3, mistralai2025magistral}. Despite its widespread adoption, the structural conditions under which reasoning distillation is actually more compute-efficient than standard IFT remain underexplored.

\paragraph{Compute-optimal scaling.}
The relationship between model capacity, dataset size, and training compute has been extensively formalized to guide the efficient pretraining of large language models \citep{kaplan2020scaling, chinchilla}. Recently, this framework has expanded to encompass inference time scaling, revealing that allocating additional compute during generation can systematically enhance reasoning capabilities \citep{snell2024scaling, brown2024largelanguagemonkeysscaling, muennighoff2025s1simpletesttimescaling}. Recent research extends this analysis to API-based models, evaluating performance as a function of token cost \citep{chen2026pricereversal} and finding analogous cost heterogeneity.

\section{Conclusion}
In this paper, we investigate whether the compute overhead of reasoning distillation is justified compared to scaling standard IFT, as a function of training and inference budgets. Surprisingly, while reasoning models boost raw performance, standard IFT lies on or near the Pareto frontier for the vast majority of configurations. Reasoning reaches the frontier only when two conditions hold simultaneously: the tasks require an open-ended output format rather than a specific knowledge domain, and the models contain at least 7B parameters. Below this capacity threshold, answer length is dictated by failure, which heavily inflates inference costs. Additionally, we demonstrate that hybrid approaches can significantly reduce these compute costs. A sequential curriculum using just 25 to 50\% reasoning data recovers most of the accuracy benefit at a fraction of the training cost, while mixed training below a 50\% corpus mix threshold improves accuracy while strictly preserving standard IFT inference efficiency.

\section*{Limitations}
This study focuses on models up to 14B parameters, covering the typical regime where distillation is utilized to enhance small model performance. While this reflects a practically motivated experimental setup, and our results establish that reasoning models capture the Pareto frontier at 7B (a finding further confirmed at the 14B scale), additional scaling experiments are necessary to verify whether the identified conditions remain stable beyond this capacity. Additionally, this work specifically examines the distillation paradigm. At the model scales we consider, reinforcement learning typically underperforms distillation and incurs substantially higher training costs, suggesting that our core observations may remain consistent in that setting. Nevertheless, a dedicated study using reinforcement learning remains a promising avenue for future work to confirm this hypothesis.

\section*{Ethics Statement}
A central motivation of this work is to better contextualize the increasing emphasis on raw accuracy in modern language model development. Recent advances have led practitioners to adopt increasingly complex training and inference strategies, often involving substantially higher computational costs, in pursuit of incremental performance gains. While such gains are valuable, they are frequently reported without a clear account of their associated training and inference efficiency trade-offs. In this paper, we aim to provide a more balanced perspective by explicitly evaluating reasoning distillation and instruction fine-tuning under matched compute budgets. By quantifying both training and inference FLOPs alongside accuracy, we hope to encourage a more cost-aware evaluation of model design choices. Our findings suggest that simpler and more efficient approaches, such as IFT or carefully designed hybrid curricula, can often match or approach the performance of more expensive reasoning-based training regimes. We do not argue against the use of compute-intensive methods when they are justified by downstream needs. However, we emphasize the importance of aligning model complexity with practical constraints and deployment requirements. We hope this work contributes to a broader shift toward more targeted model development, where improvements in accuracy are systematically weighed against their computational and environmental costs.  We additionally release all code and paired training datasets (IFT and reasoning outputs for the exact same inputs) to enable reproducibility and future controlled studies on reasoning.

\section*{Acknowledgments}
We sincerely thank the Jean Zay supercomputer (GENCI-IDRIS-CNRS) through grant AD010617149. Furthermore, we gratefully acknowledge the support of the French government through the France 2030 program as part of the ArGiMi project.

\clearpage

\bibliography{custom}

\clearpage
\appendix

\begin{figure*}[t]
\begin{equation}
\begin{aligned}
\text{FLOPs}_{\text{forward}} &=
\underbrace{2 \, l \, V \, d_\text{model}}_{\text{embeddings}} \;+\; \underbrace{ \left(
6 \, l \, d_\text{model}^2 \;+\;
2 \, l^2 \, d_\text{model} \;+\;
3 \, l^2 \, h \;+\;
2 \, l^2 \, d_\text{model} \;+\;
2 \, l \, d_\text{model}^2
\right) \cdot N_l}_{\text{attention}} \\
& \quad \;+\ \underbrace{4 \, l \, d_\text{model} \, d_\text{ff} \, N_l}_{\text{feed-forward}} \;+\;
\underbrace{2 \, l \, d_\text{model} \, V}_{\text{output logits}}
\end{aligned}
\label{eq:flops_forward}
\end{equation}
\begin{equation}
\text{FLOPs}_{\text{training step}} = 3 \cdot \text{FLOPs}_{\text{forward}}
\label{eq:training_step}
\end{equation}
\begin{equation}
\text{FLOPs}_{\text{training}} = \sum_{i=1}^{N_s} \text{FLOPs}_{\text{training step}} (i)
\label{eq:training}
\end{equation}
\end{figure*}

\begin{figure*}[t]
\begin{equation}
\begin{aligned}
\text{FLOPs}_{\text{inference}} &=
\underbrace{2 \, l_p \, d_\text{model} \, V}_{\text{embeddings}} \;+\;
\underbrace{ \left(
6 \, l_p \, d_\text{model}^2 \;+\;
2 \, l_p^2 \, d_\text{model} \;+\;
3 \, l_p^2 \, h \;+\;
2 \, l_p^2 \, d_\text{model} \;+\;
2 \, l_p \, d_\text{model}^2
\right) \cdot N_l}_{\text{attention}} \\
&\quad + \underbrace{4 \, l_p \, d_\text{model} \, d_\text{ff} \, N_l}_{\text{feed-forward}} \;+\;
\underbrace{2 \, d_\text{model} \, V}_{\text{output logits}}
\end{aligned}
\label{eq:flops_inference}
\end{equation}
\begin{equation}
\begin{aligned}
\text{FLOPs}_{\text{inference w/ cache}} &=
\underbrace{2 \, d_\text{model} \, V}_{\text{embedding}} \;+\;
\underbrace{\left(
6 \, d_\text{model}^2 \;+\;
2 \, l_p \, d_\text{model} \;+\;
3 \, l_p \, h \;+\;
2 \, l_p \, d_\text{model} \;+\;
2 \, d_\text{model}^2
\right) \cdot N_l}_{\text{attention}} \\
& \quad \;+\; \underbrace{4 \, d_\text{model} \, d_\text{ff} \, N_l}_{\text{feed-forward}} \;+\;
\underbrace{2 \, d_\text{model} \, V}_{\text{logits}}
\end{aligned}
\label{eq:flops_inference_cache}
\end{equation}
\begin{equation}
\text{FLOPs}_\text{multi-token inference} =
\text{FLOPs}_\text{inference}(l_p) +
\sum_{i=1}^{l_g} \text{FLOPs}_\text{inference w/ cache}(l_p + i)
\label{eq:flops_inference_multi}
\end{equation}
\end{figure*}

\section{Distillation Setup Positioning}
\label{apd:distillation_positioning}
While frontier laboratories increasingly rely on reinforcement learning (RL) to train their largest models, recent technical reports from DeepSeek-R1 \citep{deepseekai2025deepseekr1}, Qwen3 \citep{qwen3}, and Magistral \citep{mistralai2025magistral} consistently identify distillation as the standard approach for training smaller models within the same family. These works show that small- and mid-scale models trained solely with RL are both less performant and less sample-efficient than models distilled from stronger teachers.

In particular, DeepSeek reports that a model distilled from DeepSeek-R1 substantially outperforms its RL-only counterpart across benchmarks (\citealp{deepseekai2025deepseekr1}, Appendix F.1), achieving an average score of 71.55 compared to 58.45. Similarly, Mistral finds that a Mistral-Small-24B model distilled from Magistral-Medium reasoning traces attains a higher average score than an RL-only counterpart (60.71 vs. 58.91; \citealp{mistralai2025magistral}, Table~3). These results establish distillation not merely as a pragmatic choice for controlled experiments, but as a necessary ingredient for strong performance at smaller scales.

\section{Prompting Details}
\label{apd:prompting_details}

This section provides examples of paired IFT-reasoning data (\autoref{tab:paired_data_general}, \autoref{tab:paired_data_math}), as well as the detailed prompts used for answer generation (\autoref{tab:evaluation_prompts}) and evaluation by LLM judges (\autoref{tab:judging_prompts}).

\section{Training Hyperparameters}
\label{apd:hyperparameters}
All training runs are performed for a single epoch on an identical number of samples with a global batch size of 32 items distributed across 16 H100 GPUs. This setup ensures that the number of samples per forward and backward pass remains strictly constant between standard IFT and reasoning training. The learning rate follows a Warmup-Stable-Decay (WSD) schedule \citep{shen2024powerschedulerbatchsize} (150-step linear warmup, constant plateau, and 300-step linear decay to 10\% of the peak value), using the \texttt{AdamW\_fused} optimizer \citep{loshchilov2019decoupledweightdecayregularization}. Peak learning rates are selected via grid search over $\{2{\times}10^{-5}$, $1{\times}10^{-5}$, $7{\times}10^{-6}$, $5{\times}10^{-6}$, $3{\times}10^{-6}$, $2{\times}10^{-6}$, $1{\times}10^{-6}\}$. \autoref{tab:appendix-hparams} lists the peak learning rates for all models and data formats.

\begin{table}[h]
\centering
\small
\input{tables/appendix/hp_lr}
\caption{Peak learning rates selected for each student model and training data format.}
\label{tab:appendix-hparams}
\end{table}

\section{FLOPs Computation}
\label{apd:flops_computation}

We compute training and inference FLOPs following \citet{chinchilla}.

\subsection{Notations}

The following notation is used throughout the FLOPs computation formulas: 

\begin{itemize}[itemsep=0pt, topsep=0pt]
    \item $V$ : vocabulary size
    \item $d_\text{model}$ : hidden dimension of the model
    \item $d_\text{ff}$ : dimension of feed-forward layers
    \item $h$ : number of attention heads
    \item $N_l$ : number of transformer layers
    \item $l$ : sequence length
    \item $l_p$ : prompt length
    \item $l_g$ : generation length
    \item $N_s$ : number of training samples
\end{itemize}

\subsection{Training FLOPs}
\label{apd:training_flops_formula}

Equations \ref{eq:flops_forward} to \ref{eq:training} present the FLOPs computations associated with model training as a function of the model architecture and sequence length.

\subsection{Inference FLOPs}

Equations \ref{eq:flops_inference} to \ref{eq:flops_inference_multi} detail the FLOPs required for model inference as a function of the model architecture, input sequence length, and output sequence length.

\section{Details on Interpolation Curves}
\label{apd:pareto_interpolation}

To predict the impact of further model scaling, we fit a saturating growth function to the observed data: $f(x) = \alpha + \beta ( 1 - \exp(-\gamma x ^ \delta ))$, where $x$ denotes FLOPs and $f(x)$ gives interpolated accuracy. Parameters are constrained so that $\alpha, \beta > 0$, $\alpha + \beta$ does not exceed teacher accuracy, $\gamma > 0$, and $0 < \delta \le 1$. The function is fitted by minimizing mean absolute error.


\section{Extended Task Evaluation}

\begin{figure*}[t]
\centering
\includegraphics[width=\textwidth]{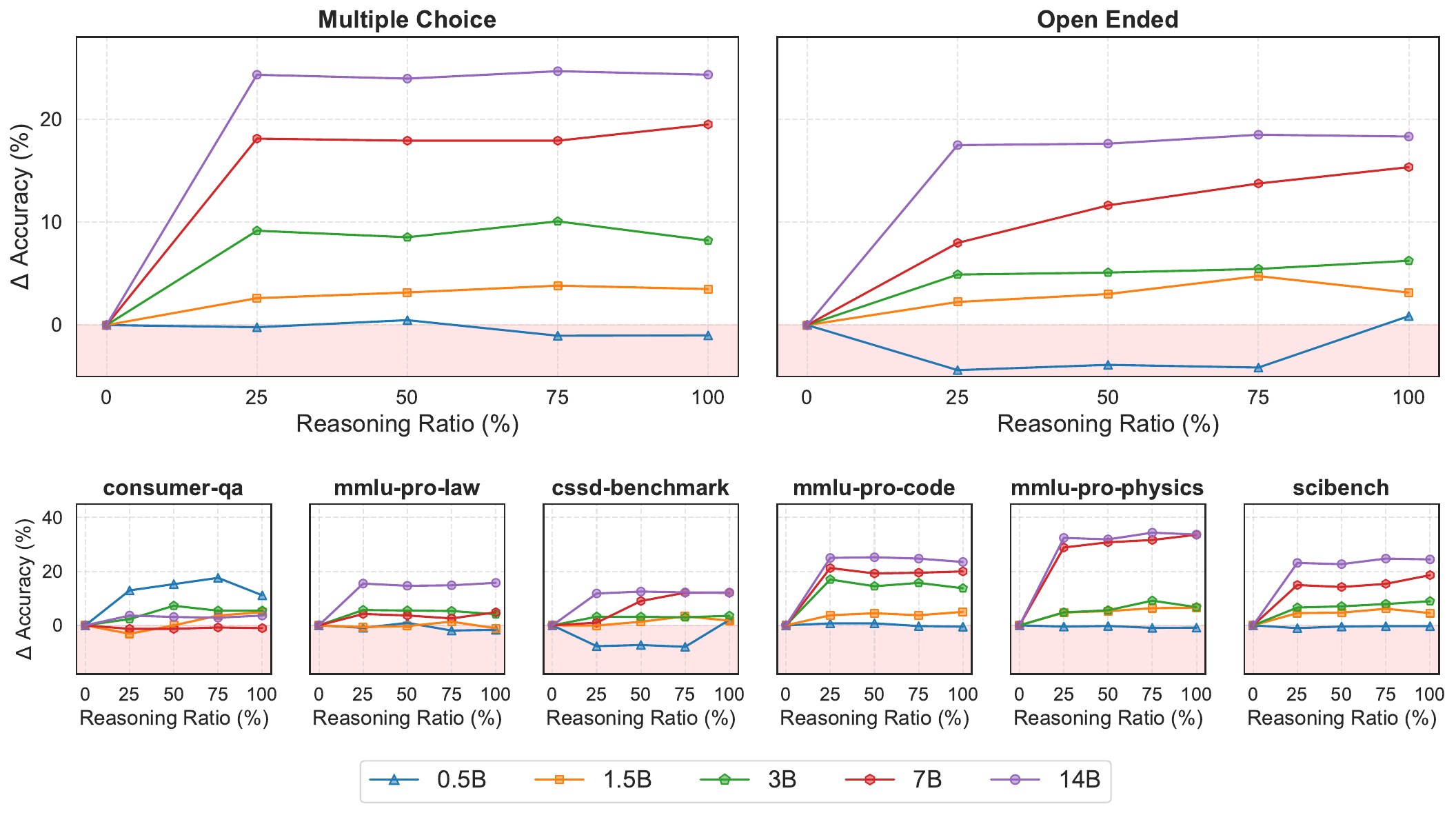}
\caption{Impact of reasoning ratio on law, code, and physics tasks in the sequential training scenario.}
\label{fig:additional_ratio}
\end{figure*}

\begin{figure*}[t]
\centering
\includegraphics[width=\textwidth]{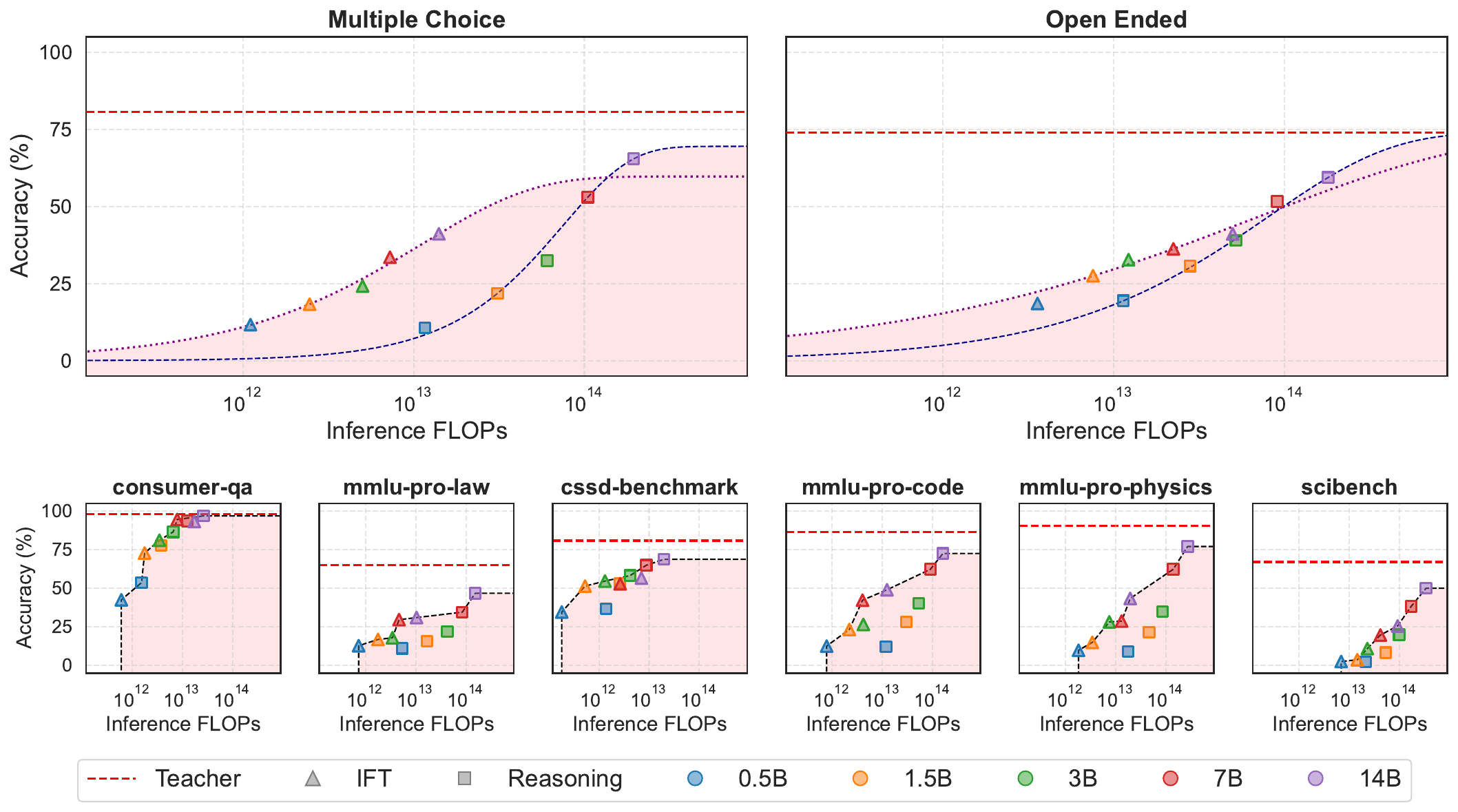}
\caption{Accuracy vs.\ inference FLOPs for law, code, and physics tasks.}
\label{fig:additional_pareto}
\end{figure*}

Main results for law, code, and physics domains are presented in \S\ref{sec:analysis} (\autoref{fig:reasoning_pareto}). Below we provide additional analyses: reasoning ratio sensitivity and inference Pareto analysis for these domains (\autoref{fig:additional_ratio}, \autoref{fig:additional_pareto}).

\section{Alternative Teacher-Student Configurations}
\label{apd:other_teacher_student}

\begin{figure*}[t]
\centering
\includegraphics[width=\textwidth]{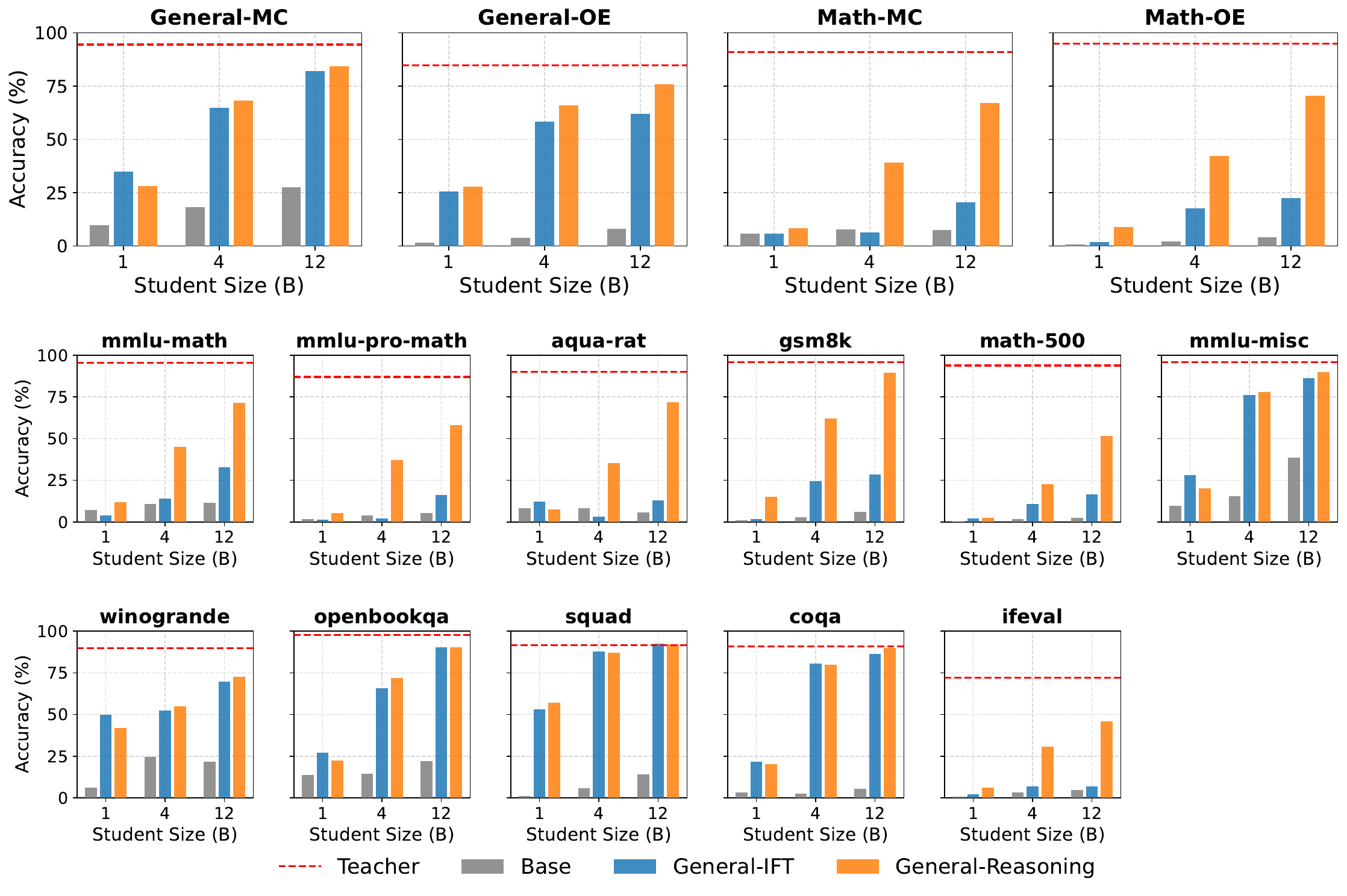}
\caption{Performance of Nemotron-Gemma teacher-student configuration.}
\label{fig:gemma_bar}
\end{figure*}

\begin{figure*}[t]
\centering
\includegraphics[width=\textwidth]{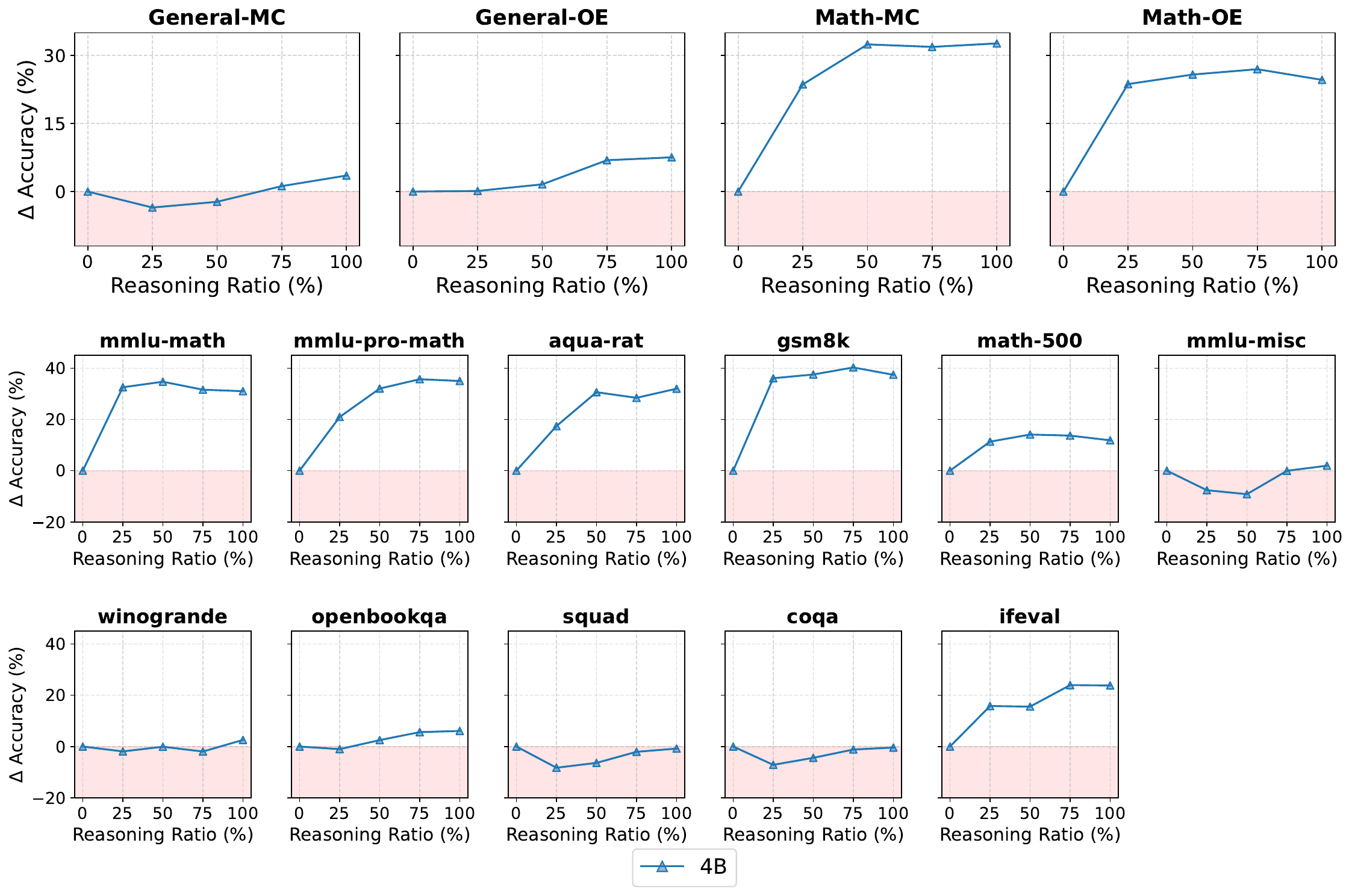}
\caption{Impact of reasoning ratio for the Nemotron-Gemma configuration at 4B.}
\label{fig:gemma_ratio}
\end{figure*}

\begin{figure*}[t]
\centering
\includegraphics[width=\textwidth]{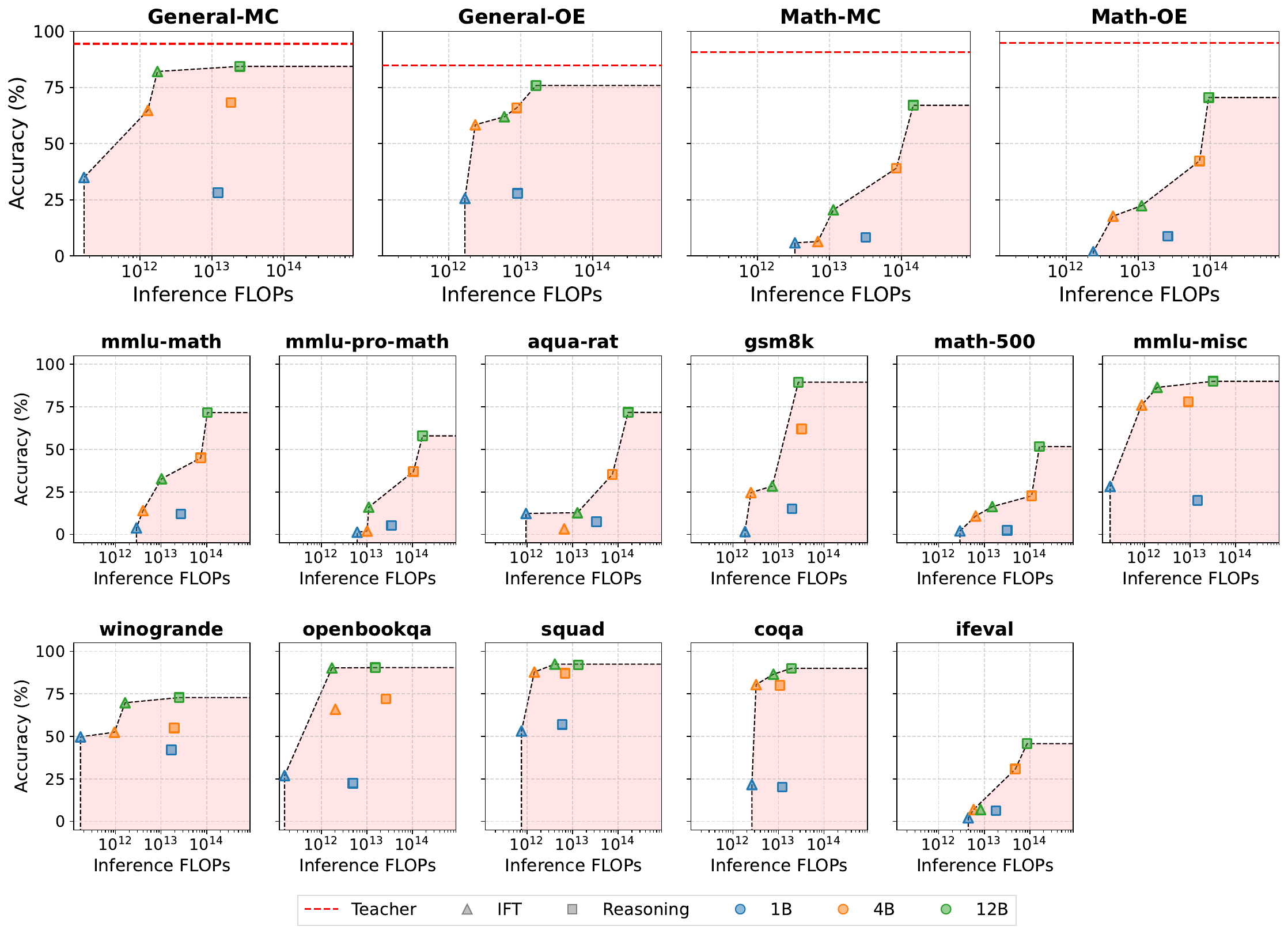}
\caption{Accuracy vs.\ inference FLOPs for the Nemotron-Gemma configuration.}
\label{fig:gemma_pareto}
\end{figure*}

We extend our study using \texttt{Nemotron-Super-49B-v1.5} as teacher with Gemma~3 students (1B, 4B, 12B), training on 200K general-domain pairs (\autoref{fig:gemma_bar}, \autoref{fig:gemma_ratio}, \autoref{fig:gemma_pareto}).

\section{Generation Early-Stopping}
\label{apd:early_stopping}

\begin{figure*}[t]
\centering
\includegraphics[width=\textwidth]{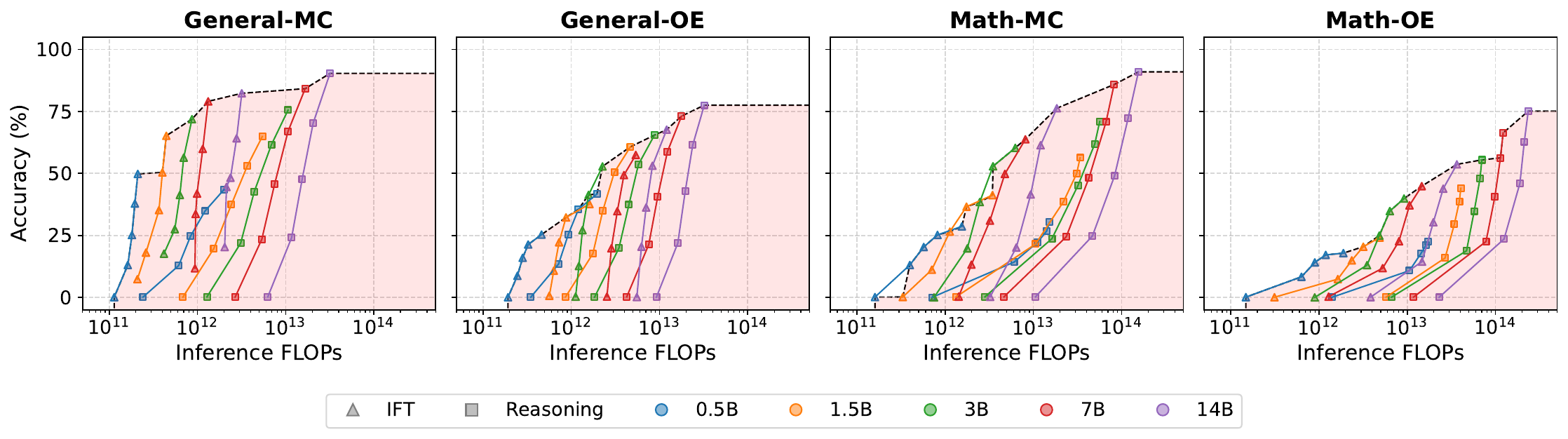}
\caption{Inference-cost impact of generation early-stopping. Each model is evaluated at five maximum-length thresholds (0th, 25th, 50th, 75th, 100th answer length percentiles).}
\label{fig:early_stopping}
\end{figure*}

\begin{figure*}[t]
\centering
\includegraphics[width=\textwidth]{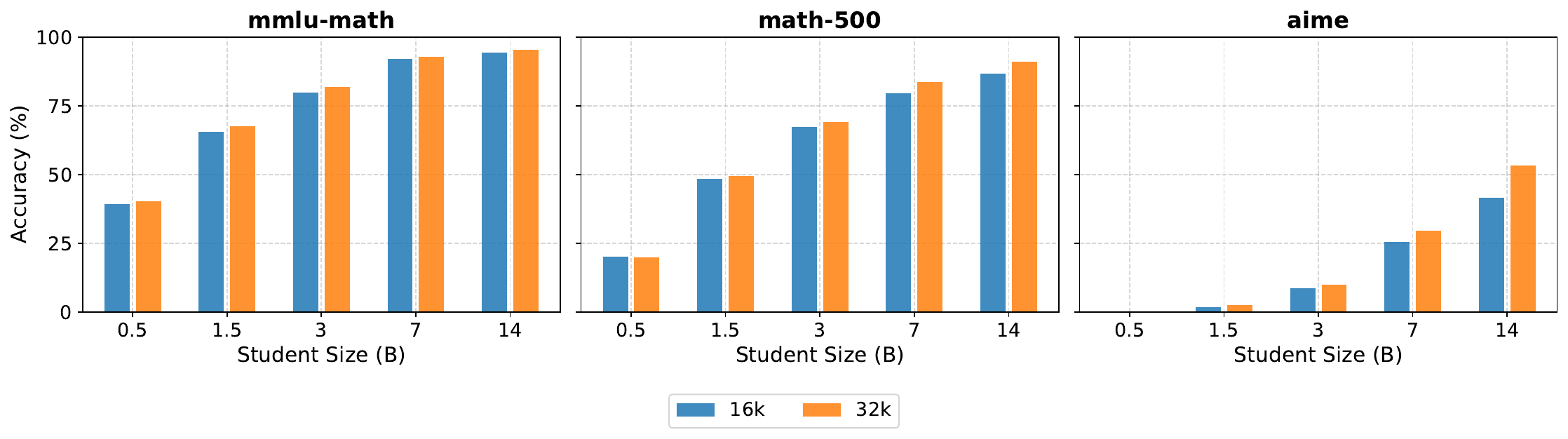}
\caption{Impact of increasing maximum generation length (16K to 32K tokens) on math tasks.}
\label{fig:high_gen_length}
\end{figure*}

Leveraging the observation that incorrect reasoning traces are typically longer, we designed a simple early stopping strategy. However, as shown in \autoref{fig:early_stopping}, this approach fails to improve the Pareto frontier, as any reduction in inference cost comes directly at the expense of accuracy. Conversely, certain mathematical tasks actively benefit from extended generation lengths in the reasoning setting (\autoref{fig:high_gen_length}), demonstrating that reasoning models are capable of extrapolating well beyond their training time trace lengths.

\section{Inference Cost Scaling Trends}
\label{apd:inference_cost_scaling_trends}

\begin{figure*}[t]
\centering
\includegraphics[width=\textwidth]{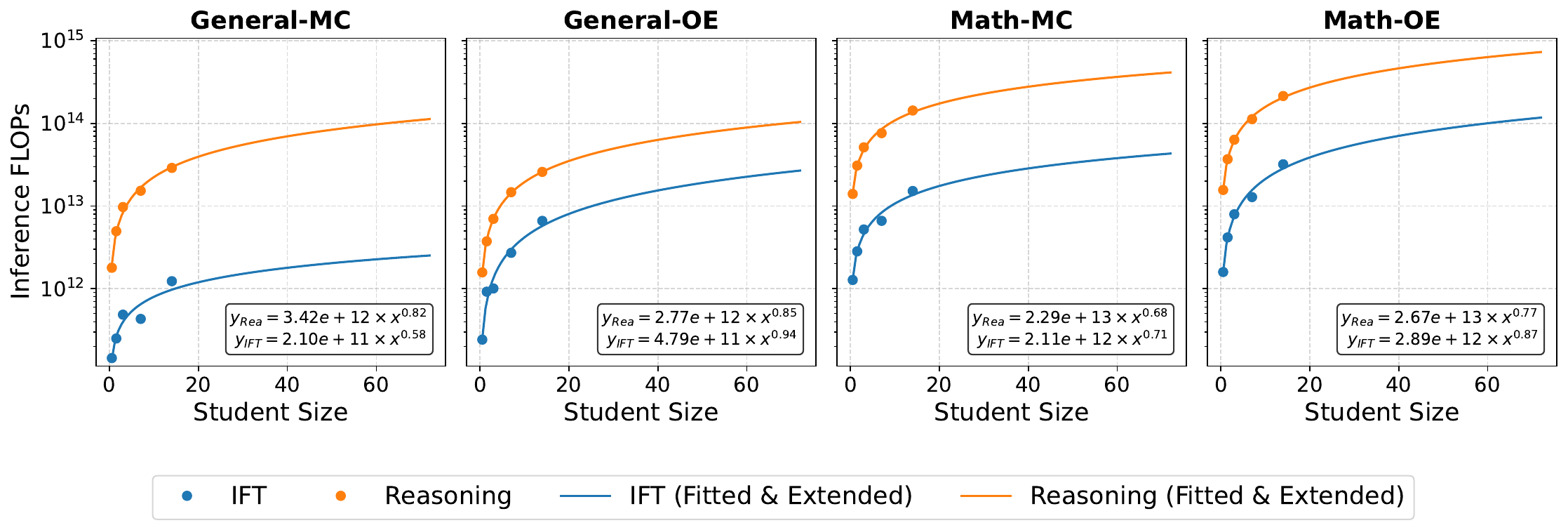}
\caption{Inference FLOPs vs.\ student model size. Power-law fits show scaling coefficients for IFT and reasoning.}
\label{fig:inference_scaling}
\end{figure*}

We fit log-linear curves to inference FLOPs as a function of model size (\autoref{fig:inference_scaling}). For General-OE, Math-MC, and Math-OE, scaling exponents are closely aligned between IFT and reasoning, consistent with reasoning answers shortening slightly faster as model size increases. For General-MC, reasoning models display larger scaling coefficients, reflecting higher cost for marginal gains.

\section{Task-Level Results}
\label{apd:task_level_results}

\begin{figure*}[t]
\centering
\includegraphics[width=\textwidth]{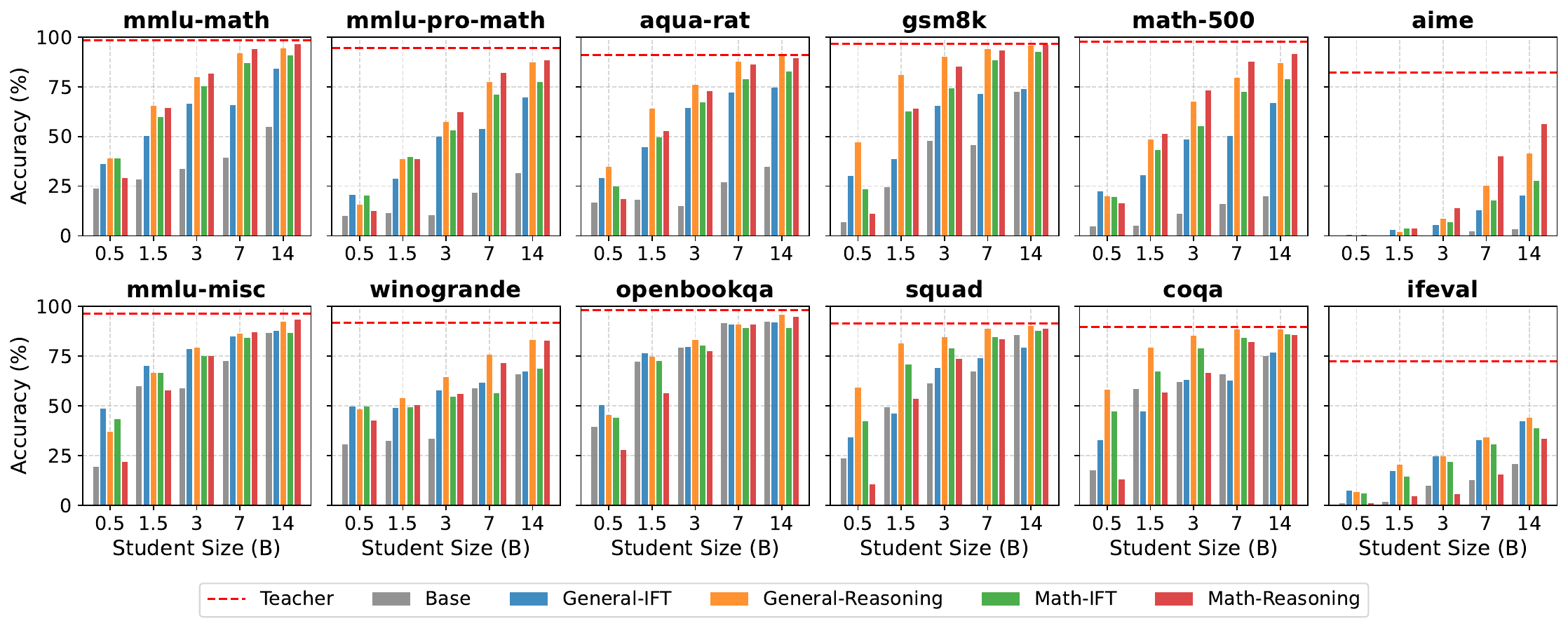}
\caption{Task-level downstream performance of mono-phasic models.}
\label{fig:results_mono_phasic_individual}
\end{figure*}

\begin{figure*}[t]
\centering
\includegraphics[width=\textwidth]{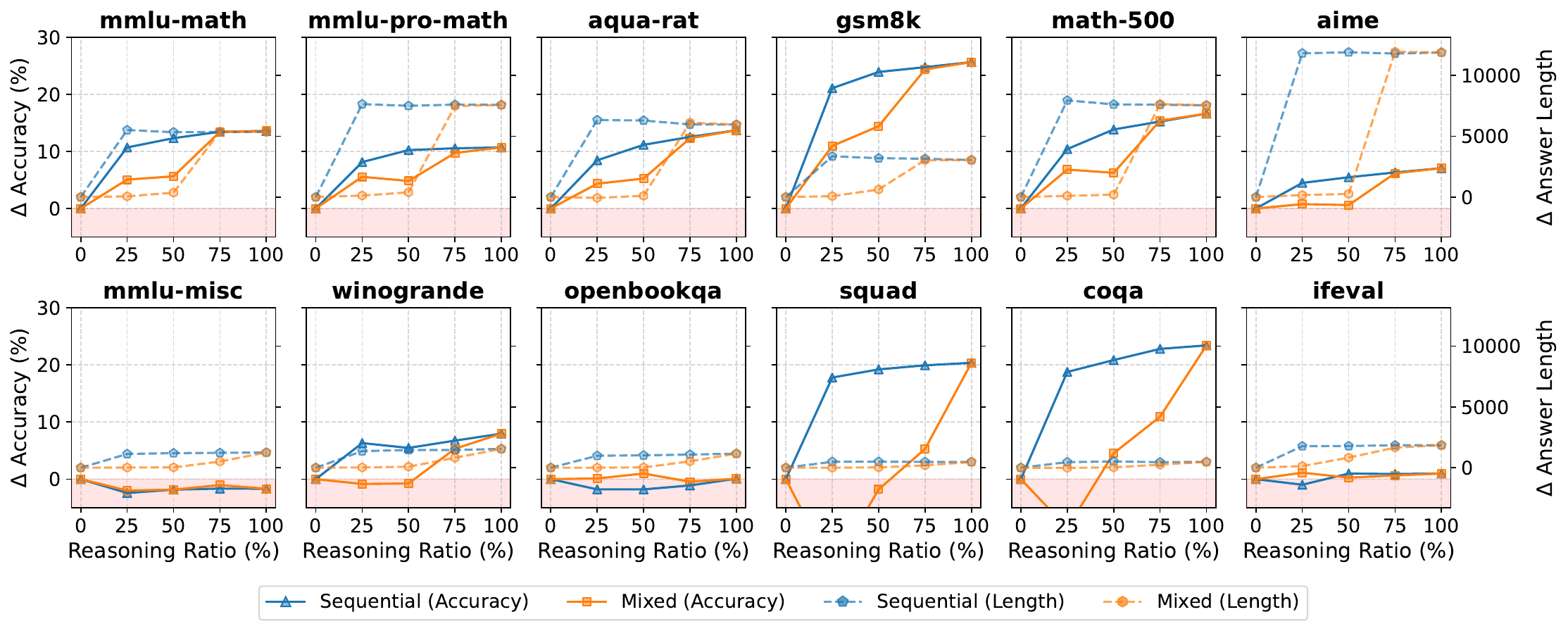}
\caption{Task-level comparison of sequential and mixed training.}
\label{fig:sequential_vs_mixed_training_individual}
\end{figure*}

\begin{figure*}[t]
\centering
\includegraphics[width=\textwidth]{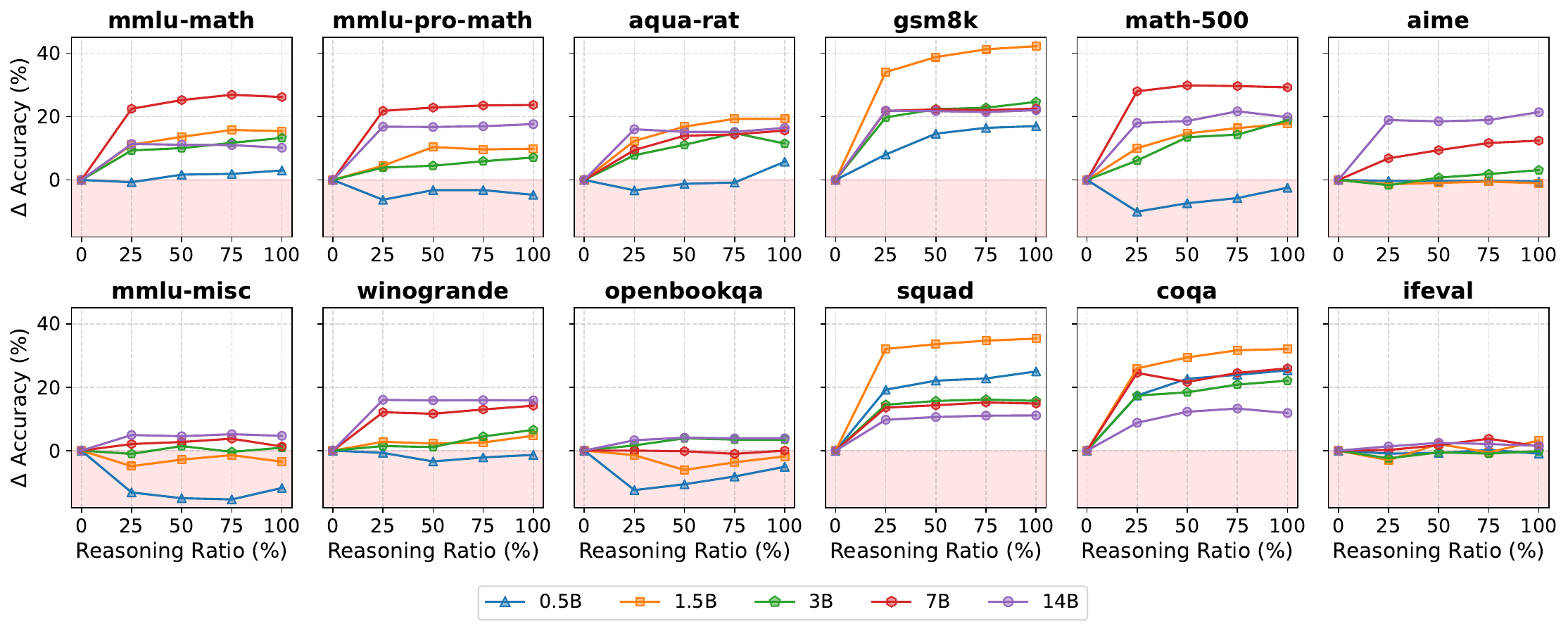}
\caption{Task-level impact of the reasoning ratio.}
\label{fig:delta_accuracy_vs_reasoning_ratio_individual}
\end{figure*}

\begin{figure*}[t]
\centering
\includegraphics[width=\textwidth]{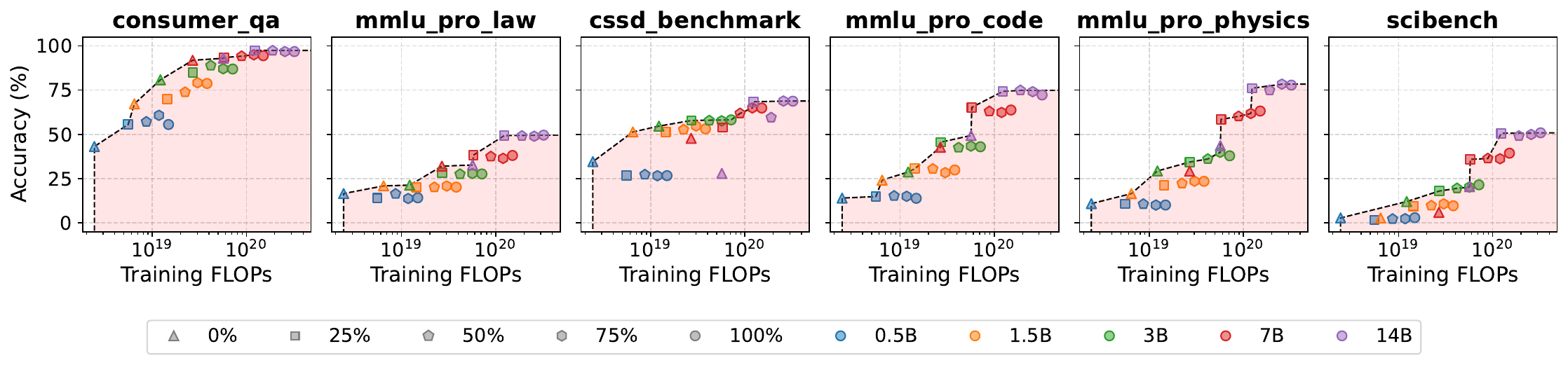}
\caption{Task-level accuracy vs.\ training FLOPs.}
\label{fig:accuracy_vs_training_flops_individual}
\end{figure*}

\begin{figure*}[t]
\centering
\includegraphics[width=\textwidth]{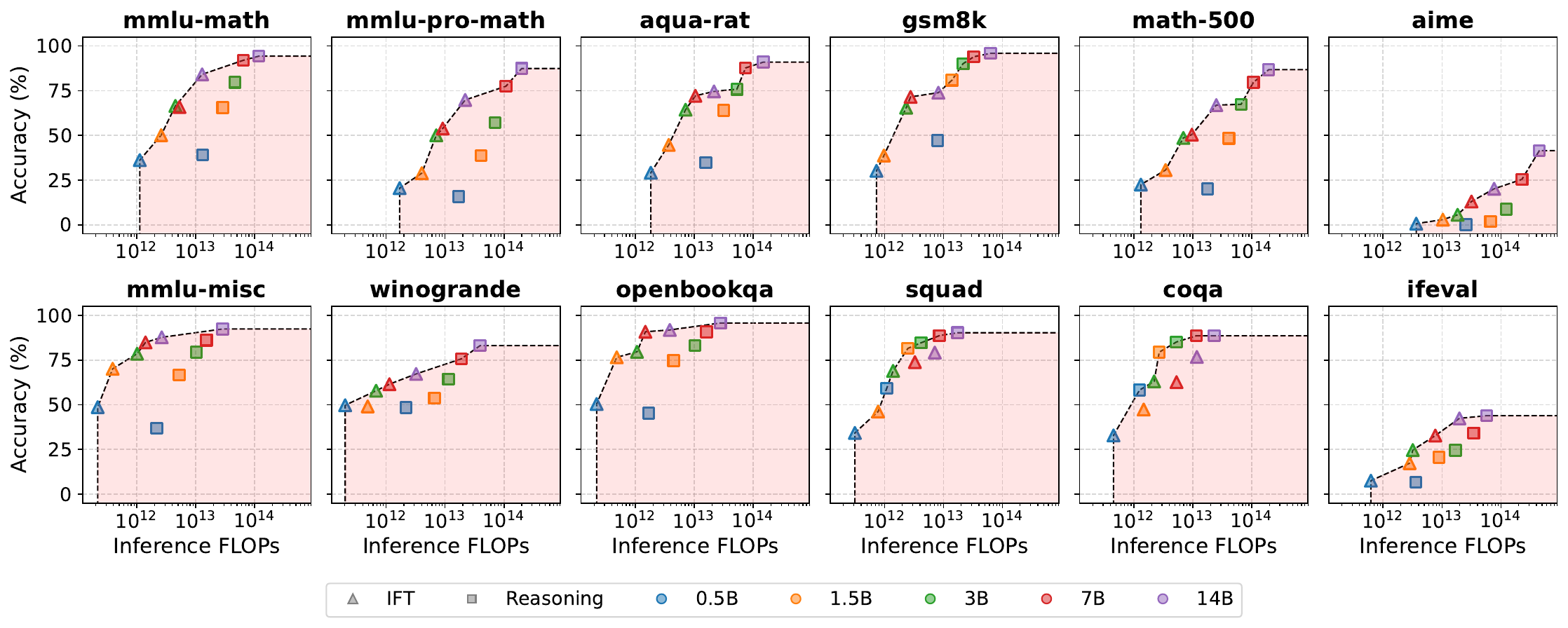}
\caption{Task-level accuracy vs.\ inference FLOPs.}
\label{fig:accuracy_vs_inference_flops_individual}
\end{figure*}

\begin{figure*}[t]
\centering
\includegraphics[width=\textwidth]{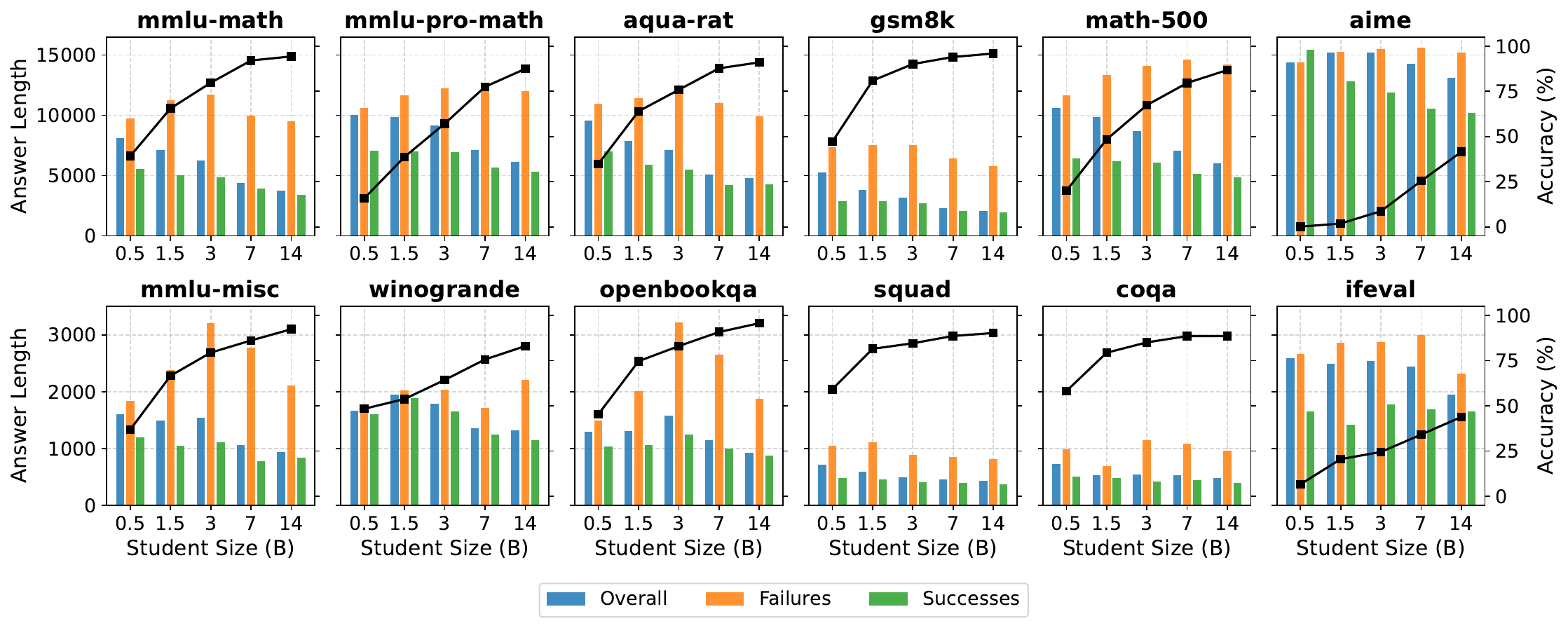}
\caption{Task-level answer length analysis.}
\label{fig:answer_length_vs_student_size_individual}
\end{figure*}

\begin{figure*}[t]
\centering
\includegraphics[width=\textwidth]{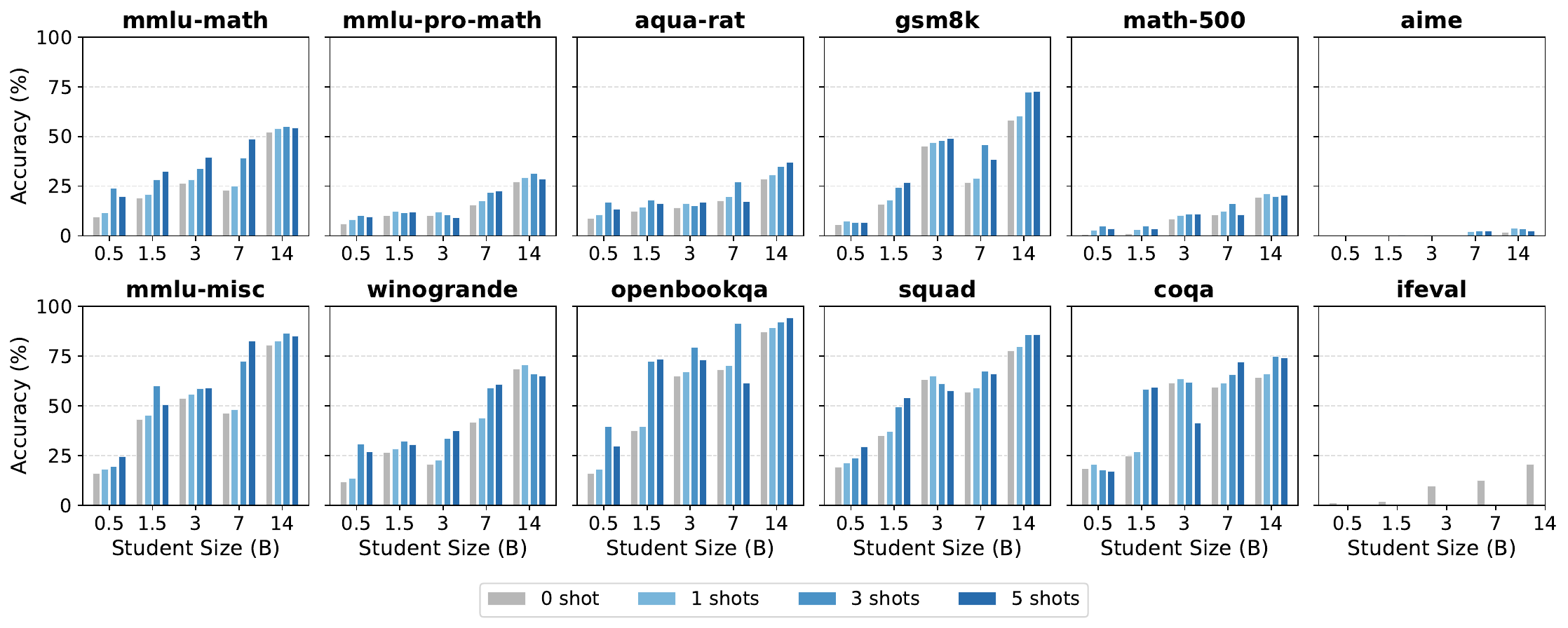}
\caption{Base student performance across 0-shot and few-shot settings.}
\label{fig:results_mono_phasic_individual_shot}
\end{figure*}

Figures \ref{fig:results_mono_phasic_individual} to \ref{fig:results_mono_phasic_individual_shot} show task-level breakdowns of the aggregated results presented in the main body.

\clearpage

\begin{table*}[t]
\centering
\input{tables/appendix/infinity_instruct_gen}
\caption{Examples of generated data from \texttt{Infinity-Instruct}, showing pairs of IFT and reasoning answers. Reasoning traces are shortened with \ldots\ due to their length.}
\label{tab:paired_data_general}
\end{table*}

\clearpage

\begin{table*}[t]
\centering
\input{tables/appendix/nemotron_postrained_gen}
\caption{Examples of generated data from \texttt{Llama-Nemotron-Post-Training-Dataset}, showing pairs of IFT and reasoning answers with character counts.}
\label{tab:paired_data_math}
\end{table*}

\clearpage

\begin{table*}[t]
\centering
\input{tables/appendix/evaluation_prompt}
\caption{Instruction prompts used for answer generation across evaluation benchmarks.}
\label{tab:evaluation_prompts}
\end{table*}

\clearpage

\begin{table*}[t]
\centering
\input{tables/appendix/validation_prompt}
\caption{Instruction prompts used for LLM-based answer assessment. Judge models follow a two-step protocol: first extracting the final answer while ignoring intermediate reasoning, then assessing semantic equivalence against the reference. We validated reliability through human verification and inter-judge agreement. Across all benchmarks, the median absolute error is 1.2, with pairwise comparisons confirming alignment: 0.9 (Nemotron vs.\ Llama), 1.0 (Nemotron vs.\ GPT-OSS), 1.8 (Llama vs.\ GPT-OSS).}
\label{tab:judging_prompts}
\end{table*}

\clearpage

\clearpage

\end{document}

%% file: tables/results_mono_phasic.tex
\begin{tabular}{ll ll ll ll ll ll c}
\toprule
\multicolumn{2}{c}{\raisebox{-1.5ex}[0pt][0pt]{\textbf{Category}}} & \multicolumn{2}{c}{\textbf{0.5B}} & \multicolumn{2}{c}{\textbf{1.5B}} & \multicolumn{2}{c}{\textbf{3B}} & \multicolumn{2}{c}{\textbf{7B}} & \multicolumn{2}{c}{\textbf{14B}} & \textbf{235B} \\
\cmidrule(lr){3-4} \cmidrule(lr){5-6} \cmidrule(lr){7-8} \cmidrule(lr){9-10} \cmidrule(lr){11-12} \cmidrule(lr){13-13}
\multicolumn{2}{c}{} & \textbf{IFT} & \textbf{Reasoning} & \textbf{IFT} & \textbf{Reasoning} & \textbf{IFT} & \textbf{Reasoning} & \textbf{IFT} & \textbf{Reasoning} & \textbf{IFT} & \textbf{Reasoning} & \textbf{Teacher} \\
\midrule
\multicolumn{13}{c}{\textit{General-purpose Training}} \\
\midrule
\raisebox{-1.5ex}[0pt][0pt]{General} & OE & 24.9 & 41.3 {\scriptsize (\up\ +16.4)} & 37.0 & 60.5 {\scriptsize (\up\ +23.5)} & 52.2 & 64.7 {\scriptsize (\up\ +12.5)} & 56.4 & 70.4 {\scriptsize (\up\ +14.0)} & 66.1 & 74.2 {\scriptsize (\up\ +8.1)} & 84.4 \\
 & MC & 49.6 & 43.5 {\scriptsize (\down\ -6.1)} & 65.2 & 65.0 {\scriptsize (\down\ -0.2)} & 72.0 & 75.6 {\scriptsize (\up\ +3.6)} & 79.1 & 84.2 {\scriptsize (\up\ +5.2)} & 82.3 & 90.4 {\scriptsize (\up\ +8.1)} & 95.4 \\
\noalign{\vspace{4pt}}
\cdashline{1-13}[1pt/2pt]
\noalign{\vspace{4pt}}
\raisebox{-1.5ex}[0pt][0pt]{Math} & OE & 17.7 & 22.4 {\scriptsize (\up\ +4.7)} & 24.1 & 43.7 {\scriptsize (\up\ +19.6)} & 39.9 & 55.4 {\scriptsize (\up\ +15.5)} & 45.0 & 66.3 {\scriptsize (\up\ +21.3)} & 53.6 & 74.7 {\scriptsize (\up\ +21.1)} & 92.1 \\
 & MC & 28.6 & 29.9 {\scriptsize (\up\ +1.3)} & 41.2 & 56.1 {\scriptsize (\up\ +14.8)} & 60.3 & 70.9 {\scriptsize (\up\ +10.6)} & 63.9 & 85.7 {\scriptsize (\up\ +21.8)} & 76.2 & 90.9 {\scriptsize (\up\ +14.7)} & 94.6 \\
\midrule
\multicolumn{13}{c}{\textit{Math-centric Training}} \\
\midrule
\raisebox{-1.5ex}[0pt][0pt]{General} & OE & 31.8 & 8.2 {\scriptsize (\down\ -23.6)} & 50.8 & 38.3 {\scriptsize (\down\ -12.5)} & 59.8 & 48.6 {\scriptsize (\down\ -11.2)} & 66.5 & 60.4 {\scriptsize (\down\ -6.1)} & 70.8 & 69.2 {\scriptsize (\down\ -1.6)} & 84.4 \\
 & MC & 45.7 & 30.8 {\scriptsize (\down\ -14.9)} & 62.9 & 54.8 {\scriptsize (\down\ -8.1)} & 70.0 & 69.5 {\scriptsize (\down\ -0.5)} & 76.5 & 83.2 {\scriptsize (\up\ +6.6)} & 81.5 & 90.3 {\scriptsize (\up\ +8.7)} & 95.4 \\
\noalign{\vspace{4pt}}
\cdashline{1-13}[1pt/2pt]
\noalign{\vspace{4pt}}
\raisebox{-1.5ex}[0pt][0pt]{Math} & OE & 14.5 & 9.2 {\scriptsize (\down\ -5.3)} & 36.5 & 39.7 {\scriptsize (\up\ +3.2)} & 45.5 & 57.5 {\scriptsize (\up\ +12.0)} & 59.6 & 73.6 {\scriptsize (\up\ +14.0)} & 66.3 & 81.3 {\scriptsize (\up\ +14.9)} & 92.1 \\
 & MC & 28.2 & 20.0 {\scriptsize (\down\ -8.2)} & 49.7 & 51.9 {\scriptsize (\up\ +2.2)} & 65.2 & 72.2 {\scriptsize (\up\ +7.0)} & 78.9 & 87.3 {\scriptsize (\up\ +8.4)} & 83.8 & 91.4 {\scriptsize (\up\ +7.6)} & 94.6 \\
\bottomrule
\end{tabular}

%% file: tables/appendix/hp_lr.tex
\begin{tabular}{lcccc}
\toprule
Model & Reasoning & IFT \\
\midrule
\texttt{Qwen2.5-0.5B} & 2e-5 & 1e-5 \\
\texttt{Qwen2.5-1.5B} & 1e-5 & 7e-6 \\
\texttt{Qwen2.5-3B}   & 7e-6 & 5e-6 \\
\texttt{Qwen2.5-7B}   & 5e-6 & 3e-6 \\
\texttt{Qwen2.5-14B}  & 3e-6 & 1e-6 \\
\midrule
\texttt{Gemma3-1B}  & 2e-5 & 1e-5 \\
\texttt{Gemma3-4B}  & 5e-6 & 3e-6 \\
\texttt{Gemma3-12B} & 2e-6 & 1e-6 \\
\bottomrule
\end{tabular}

%% file: tables/appendix/infinity_instruct_gen.tex
\begin{tabular}{|p{0.30\textwidth}|p{0.70\textwidth}|}
\hline
\textbf{Question} & \textbf{Answer} \\ \hline
\vspace{1pt}
Lotto 6/49 is one of three national lottery games in Canada. Launched on June 12, 1982, Lotto 6/49 was the first nationwide Canadian lottery game to allow players to choose their own numbers. Previous national games, such as the Olympic Lottery, Loto Canada and Superloto used pre-printed numbers on tickets. Lotto 6/49 led to the gradual phase-out of that type of lottery game in Canada. Based on the paragraph above can we conclude that Canada's Lottery 6/49 was very popular in the 70s? 
&
\begin{minipage}[t]{\hsize}
\vspace{1pt}
\textbf{IFT:} No, we cannot conclude that "Canada's Lottery 6/49 was very popular in the 70s" based on the paragraph. The text states that Lotto 6/49 was launched on June 12, 1982, which is after the 1970s, so the statement is incorrect.

\noindent\rule{\hsize}{0.4pt}\\

\textbf{Reasoning:} \texttt{<think>} Okay, let's see. The user is asking if we can conclude that Canada's Lotto 6/49 was very popular in the 70s based on the given paragraph. First, I need to check the dates mentioned.  

\begin{center}
\ldots
\end{center}

The answer should state that the conclusion is incorrect because the game started in 1982, which is after the 70s.\texttt{</think>}\\
No, we cannot conclude that "Canada's Lottery 6/49 was very popular in the 70s" based on the paragraph. The text explicitly states that Lotto 6/49 was \textbf{launched on June 12, 1982}, which is in the early 1980s, not the 1970s. The paragraph also notes that earlier national lottery games (e.g., Olympic Lottery, Loto Canada, Superloto) used pre-printed numbers and were phased out after Lotto 6/49's introduction. There is no information provided about the popularity of any lottery games in the 1970s, nor was Lotto 6/49 even in existence during that decade. The conclusion is therefore unsupported.\\
\end{minipage} \\ \hline
\vspace{1pt}
I know that the answer to the question What are isolators found in microwave antennas made of? is in A necessary condition for the aforementioned reciprocity property is that the materials in the antenna and transmission medium are linear and reciprocal. Most materials used in antennas meet these conditions, but some microwave antennas use high-tech components such as isolators and circulators, made of nonreciprocal materials such as ferrite. These can be used to give the antenna a different behavior on receiving than it has on transmitting, which can be useful in applications like radar. Can you tell me what it is?
&
\begin{minipage}[t]{\hsize}
\vspace{1pt}
\textbf{IFT:} The isolators found in microwave antennas are made of \textbf{nonreciprocal materials such as ferrite}.

\noindent\rule{\hsize}{0.4pt}\\

\textbf{Reasoning:} \texttt{<think>} Okay, let's see. The user is asking about what isolators found in microwave antennas are made of. They provided a passage that should contain the answer. First, I need to read through the passage carefully. The key part here is where it mentions isolators and circulators. The sentence says, "made of nonreciprocal materials such as ferrite." So the answer should be ferrite. Wait, but let me make sure. The passage starts by talking about reciprocity in antennas, but then introduces isolators and circulators as exceptions. It says these components are made of nonreciprocal materials. The example given is ferrite. So yes, the answer is definitely ferrite. I should check if there are any other materials mentioned, but the text only lists ferrite as an example. The user might be looking for that specific term. No other materials are named here. So the answer should be ferrite.\texttt{</think>}\\
Answer: The isolators found in microwave antennas are made of \textbf{ferrite}, a nonreciprocal material.\\
\end{minipage} \\ \hline

\end{tabular}

%% file: tables/appendix/nemotron_postrained_gen.tex
\begin{tabular}{|p{0.30\textwidth}|p{0.70\textwidth}|}
\hline
\textbf{Question} & \textbf{Answer} \\ \hline
\vspace{1pt}
Solve the following math problem. Make sure to put the answer (and only the answer) inside \verb|\boxed{}|.
Evaluate the limit
\[
\lim_{n \to \infty} \left(\frac{n-10}{n+1}\right)^{3n+1}
\]
&
\begin{minipage}[t]{\hsize}
\vspace{1pt}
\textbf{IFT:} We are asked to evaluate the limit: \( \lim_{n \to \infty} \left( \frac{n-10}{n+1} \right)^{3n+1} \). Step 1: Simplify the base expression \(\frac{n - 10}{n + 1} = \frac{n\left(1 - \tfrac{10}{n}\right)}{n\left(1 + \tfrac{1}{n}\right)} = \frac{1 - \tfrac{10}{n}}{1 + \tfrac{1}{n}}\).  

\begin{center}
\ldots
\end{center}

\((3 + \tfrac{1}{n}) = -33\). Now exponentiate: \( L = e^{-33} \). Final Answer: \verb|\boxed{e^{-33}}|\\[4pt]

\textbf{Total characters:} 1,417

\noindent\rule{\hsize}{0.4pt}\\

\textbf{Reasoning:} \texttt{<think>} Okay, so I need to evaluate the limit as \(n \to \infty\) of \(\bigl(\tfrac{n - 10}{n + 1}\bigr)^{3n + 1}\). Hmm, let's see. I remember that limits involving expressions raised to variables can sometimes be solved using the exponential function and  

\begin{center}
\ldots
\end{center}

Thus: \(\ln L \approx \lim_{n \to \infty} (3n + 1) \cdot \left(-\tfrac{11}{n + 1}\right) = -11 \cdot \lim_{n \to \infty} \tfrac{3n + 1}{n + 1} = -11 \cdot 3 = -33\). Exponentiating both sides: \( L = e^{-33} \). Final Answer: \verb|\boxed{e^{-33}}|\\

\textbf{Total characters:} 11,225\\
\end{minipage} \\ \hline
\vspace{1pt}
Solve the following math problem. Make sure to put the answer (and only the answer) inside \verb|\boxed{}|.
Find the number of integer solutions to the equation \( xyz = 2003 \).
&
\begin{minipage}[t]{\hsize}
\vspace{1pt}
\textbf{IFT:} We are asked to find the number of integer solutions to the equation \( xyz = 2003 \). Step 1: Prime Factorization. First, factor 2003: \(2003 \div 17 = 117.823 \quad \text{(not divisible by 17)}\).  

\begin{center}
\ldots
\end{center}

Try dividing by small primes up to \(\sqrt{2003} \approx 44.76\).  

Eventually, we find: \verb|\boxed{12}|.\\

\textbf{Total characters:} 2,806\\

\noindent\rule{\hsize}{0.4pt}\\

\textbf{Reasoning:} \texttt{<think>} Okay, so I need to find the number of integer solutions to the equation \( xyz = 2003 \). Hmm, let me start by understanding what exactly the problem is asking for. We need all ordered triples of integers \((x, y, z)\) such that when you multiply them together, you get 2003.

\begin{center}
\ldots
\end{center}

Thus, only the above 12 combinations are valid. Final Answer: \verb|\boxed{12}|.\\

\textbf{Total characters:} 26,747\\
\end{minipage} \\ \hline

\end{tabular}

%% file: tables/appendix/evaluation_prompt.tex
\begin{tabularx}{\textwidth}{| >{\RaggedRight}m{0.2\textwidth} | >{\RaggedRight}X |}
\hline
\textbf{Benchmark} & \textbf{Instruction} \\
\hline
\texttt{gsm8k} & Solve the following math problem. Make sure to put the answer (and only answer) inside \textbackslash boxed\{\}. \\
\hline
\texttt{math\_500} & Solve the following math problem. Make sure to put the answer (and only answer) inside \textbackslash boxed\{\}. \\
\hline
\texttt{aime} & Solve the following math problem. Make sure to put the answer (and only answer) inside \textbackslash boxed\{\}. \\
\hline
\texttt{mmlu\_math} & Solve the following math problem. Make sure to put the answer (and only answer) inside \textbackslash boxed\{\}. \\
\hline
\texttt{mmlu\_pro\_math} & Solve the following math problem. Make sure to put the answer (and only answer) inside \textbackslash boxed\{\}. \\
\hline
\texttt{aqua\_rat} & Solve the following math problem. Make sure to put the answer (and only answer) inside \textbackslash boxed\{\}. \\
\hline
\texttt{winogrande} & Given a sentence with a blank (\_) and two possible options, choose the option that correctly fills the blank so that the sentence makes the most logical sense. Make sure to put the answer (and only answer) inside \textbackslash boxed\{\}. \\
\hline
\texttt{openbookqa} & Select the option that best completes the scenario based on everyday reasoning about cause and effect. Make sure to put the answer (and only answer) inside \textbackslash boxed\{\}. \\
\hline
\texttt{squad} & Read the passage and answer the question by selecting the text span from the passage that best answers it. Make sure to put the answer (and only answer) inside \textbackslash boxed\{\}. \\
\hline
\texttt{mmlu\_misc} & Answer the following multiple-choice question by selecting the option that best fits the correct knowledge. Make sure to put the answer (and only answer) inside \textbackslash boxed\{\}. \\
\hline
\texttt{coqa} & Read the passage and answer the question by selecting the text span from the passage that best answers it. Make sure to put the answer (and only answer) inside \textbackslash boxed\{\}. \\
\hline
\texttt{ifeval} & Answer the following instruction. \\
\hline
\end{tabularx}

%% file: tables/appendix/validation_prompt.tex
\begin{tabular}{|l|p{0.8\textwidth}|}
\hline
\textbf{Benchmark} & \textbf{Instruction} \\
\hline
Default & You will be given a Question, a User Answer (only its ending is shown due to length), and a Ground Truth.\newline
Your task is not to answer the question, but to say if the user answer is equivalent in meaning to the ground truth.\newline
\newline
First, extract the final result from both the User Answer and the Ground Truth Answer, based on the Question.\newline
Then, compare the two final results and determine whether they convey the same meaning.\newline
If they are equivalent, respond with \textbackslash boxed\{yes\}.\newline
If they are not equivalent, or if the User Answer does not contain a valid answer, respond with \textbackslash boxed\{no\}.\newline
\newline
Question:\newline
\{question\}\newline
\newline
User Answer:\newline
\{answer\}\newline
\newline
Ground Truth:\newline
\{truth\} \\
\hline
\texttt{ifeval} & You will be given an Instruction and a User Answer (only its ending is shown due to length).\newline
Your task is not to answer the Instruction, but to determine whether the User Answer follows all the formal requirements stated in the Instruction. If the User Answer contains a thinking process, you should ignore it and only focus on the final answer.\newline
\newline
First, identify every explicit requirement in the Instruction (e.g., no commas, maximum word count, required word occurrences, formatting rules).\newline
Then, compare the User Answer against these requirements.\newline
If all requirements are satisfied, respond with \textbackslash boxed\{yes\}.\newline
If any requirement is violated, respond with \textbackslash boxed\{no\}.\newline
\newline
Question:\newline
\{question\}\newline
\newline
User Answer:\newline
\{answer\} \\
\hline
\end{tabular}